\documentclass[11pt]{article}

\usepackage[preprint]{acl}

\usepackage{times}
\usepackage{latexsym}

\usepackage[T1]{fontenc}

\usepackage[utf8]{inputenc}

\usepackage{microtype}

\usepackage{inconsolata}

\usepackage{graphicx}

\usepackage{amssymb}
\usepackage{amsmath}
\usepackage{bbm}

\usepackage{url}
\usepackage[skins,breakable]{tcolorbox}
\usepackage{booktabs}
\usepackage{tabularx}
\usepackage{longtable}
 \usepackage{arydshln}

\usepackage{xcolor}

\definecolor{textgray}{RGB}{50,50,50}
\definecolor{bordercolor}{RGB}{180,180,180}

\definecolor{pipeline_decomp}{HTML}{00994D}
\definecolor{pipeline_harm}{HTML}{0066CC}
\definecolor{pipeline_verify}{HTML}{990099}
\definecolor{pipeline_score}{HTML}{FF8000}

\newcommand{\badge}[2][red]{%
  \tikz[baseline={([yshift=-1pt]b.base)}]
    \node[circle, fill=#1, text=white, inner sep=1pt,
    minimum size=1.0em, font=\bfseries\scriptsize] (b) {#2};%
}

\title{\textsc{VetScore}: Risk-Weighted Fact Verification for Veterinary Long-Form~QA with Citations}

\author{
 \textbf{Ivan Kartáč}$^{1,2}$ \quad \textbf{Jan Tovarys}$^{1}$ \quad \textbf{Mateusz Lango}$^{2}$ \quad \textbf{Ondřej Dušek}$^{2}$ \vspace{0.5em}
 \\
 \textsuperscript{1}PrimVeterinary
 \\
 \textsuperscript{2}Charles University, Faculty of Mathematics and Physics
\\
 \small{
   \textbf{Correspondence:} \href{kartac@ufal.mff.cuni.cz}{kartac@ufal.mff.cuni.cz}, \href{jenda@primveterinary.com}{jenda@primveterinary.com}
 }
}

\begin{document}
\maketitle
\begin{abstract}
Citation excerpts can be used to increase the reliability of generated outputs and their faithfulness to cited sources, which is especially important in high-stakes domains such as human and veterinary medicine. However, this does not guarantee that generated claims are faithful to the provided excerpts. We present \textsc{VetScore}, a multi-step evaluation method for veterinary long-form question answering, designed to assess how well generated claims are supported by the provided excerpts, weighing this information by each claim's harm potential. \textsc{VetScore} first segments the output and decomposes it into individual claims, then scores each claim with respect to its harm potential, evaluates its faithfulness to source excerpts, and finally calculates the overall risk-adjusted score. We collect an expert-annotated meta-evaluation dataset and apply an item response theory model to analyze the annotators' sensitivity to harm potential and to aggregate their annotations. Evaluating our approach with a range of judge models, we show that it achieves high correlations with veterinary experts while offering explainability across multiple dimensions.\footnote{Code and data are available at \url{https://github.com/PrimVet/VetScore}}
\end{abstract}

\section{Introduction}
\label{sec:introduction}

\begin{figure}[t!]
    \centering
    \includegraphics[width=\columnwidth]{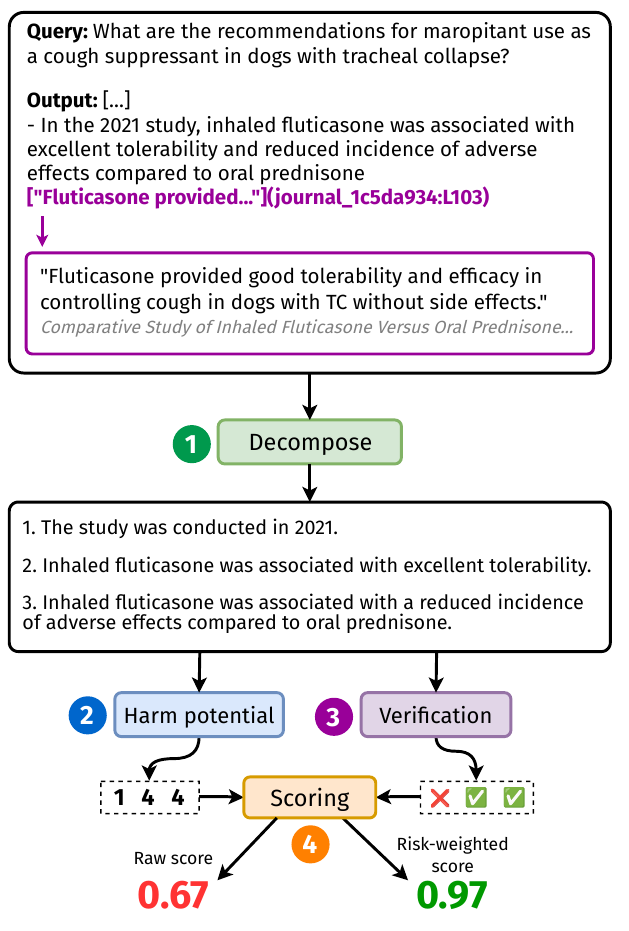}
    \caption{Overview of the \textsc{VetScore} pipeline. \badge[pipeline_decomp]{1} \textbf{Claim decomposition} splits a segment of the output into atomic claims, \badge[pipeline_harm]{2} \textbf{Harm potential scoring} assigns a score (1-5) to each claim, indicating the potential harm of the claim, \badge[pipeline_verify]{3} \textbf{Fact verification} decides for each claim whether it is supported by the given source excerpt(s), and \badge[pipeline_score]{4} \textbf{Combined score} aggregates verification and harm potential scores to a final score for the segment.}
    \label{fig:diagram}
\end{figure}

While large language models (LLMs) are being increasingly applied in the medical and veterinary domains \citep{thirunavukarasu2023large,singhal2025toward,shah2026conversationalmedicalaieyes,chu2024chatgpt}, they have been shown to produce potentially harmful outputs \citep{wu2025firstnoharmclinicallysafe}. Recent work in NLP has called for attributable text generation to make outputs more reliable and trustworthy through including citations \cite{rashkin-etal-2023-measuring,gao-etal-2023-enabling,huang-chang-2024-citation}. Although a number of attributable generation methods have been proposed \cite{nakano2022webgptbrowserassistedquestionansweringhuman,menick2022teachinglanguagemodelssupport}, including citations in the outputs does not guarantee sufficient factual accuracy or faithfulness to the cited sources \cite{wallat2024correctnessfaithfulnessragattributions,shi2026citeauditciteditdid,onweller2026citedverifiedparsingevaluating}. Moreover, citations are often present only in the form of URLs or general references to the cited sources, which offloads the cognitive demand of the verification on the user. Although some methods address this by providing source excerpts or passages along with their citations \cite{nakano2022webgptbrowserassistedquestionansweringhuman,menick2022teachinglanguagemodelssupport,gao-etal-2023-enabling}, there are two main reasons why we still need to be able to reliably verify these citations at scale: (1) claims contained in the LLM-generated outputs are not necessarily faithful to the provided excerpts \cite{liu-etal-2023-evaluating,gao-etal-2023-enabling,wallat2024correctnessfaithfulnessragattributions}, and (2) citations can increase over-reliance, where users become less critical of the attributed outputs, regardless of citation correctness \citep{ding2025citations,li2025humantrustaisearch}. These limitations make reliable and informative automatic verification of factual claims and their faithfulness to the cited source excerpts a critical component of an evidence-based generation system. This is particularly important if the provided claims inform clinical reasoning and decisions.

There is a line of related work on factuality evaluation, often based on the \emph{decompose-then-verify} paradigm \cite{min-etal-2023-factscore,wei2024long}. These approaches address the problem by first decomposing the outputs into individual atomic claims, followed by verification of each claim against the retrieved sources.
However, they typically treat each claim equally, while in human or veterinary medicine, we need to distinguish degrees of potential harm present in the provided information and penalize unverified high-risk claims (such as claims about drug dosage) more than low-risk claims (such as meta-comments about a year of a cited study).

In this paper, we present \textsc{VetScore}, a fact verification pipeline designed to evaluate faithfulness of outputs to their cited excerpts in long-form evidence-based veterinary QA. \textsc{VetScore} is based on the decompose-then-verify paradigm, extended with risk-adjusted scoring where each claim is weighted by its harm potential. Unlike existing approaches, this weighting makes the final verification scores clinically more informative and allows for fine-grained validation based on severity. Specifically, \textsc{VetScore} scores each claim for its harm potential and independently decides whether the claim is supported by the provided excerpts. These two signals are then aggregated into a final score for the given output segment (Figure~\ref{fig:diagram}). This pipeline can be used both for output evaluation and real-time validation, while the fine-grained harm potential scoring allows making decisions based on thresholds, e.g., flagging parts of the output with unverified high-risk claims. \textsc{VetScore} also provides fine-grained explainability through decomposition into claims, modularity, and concise natural language explanations.

Our contributions are as follows:

\begin{itemize}
    \item We present \textsc{VetScore}, an efficient and explainable modular verification pipeline providing risk-weighted faithfulness scores for veterinary long-form QA with citation excerpts.
    \item Our approach is tested with 12 different LLMs and validated by veterinary experts (including students and professionals), achieving high correlations with human judgments.
    \item We perform ablation experiments, analysis of scoring distributions, consistency across runs, and an item response theory modeling of annotator behavior.
\end{itemize}

\section{Related Work}
\label{sec:related_work}

\paragraph{Attributable text generation} While a number of different approaches to attributable text generation exist, \emph{post-retrieval} generation, where external information is retrieved before generating the output, is the most common \citep{schreieder2025attributioncitationquotationsurvey}. RAG \citep{lewis2020retrieval} is the most prominent of these methods, but needs to be extended to include citations. \citet{nakano2022webgptbrowserassistedquestionansweringhuman} trained an LLM to generate answers with citations, while \citet{menick2022teachinglanguagemodelssupport} also include source excerpts in the outputs. \citet{huang-etal-2024-learning} first extract excerpts from retrieved sources, and then condition the generation on these selected excerpts. \citet{zhang-etal-2025-longcite} use a different approach, generating only pointers to retrieved sentences, which can effectively serve as inline quotes. Source excerpts are also incorporated into the outputs of recent systems for automatic scientific literature synthesis such as OpenScholar \citep{asai2026synthesizing}. ALCE \citep{gao-etal-2023-enabling} is the first benchmark for long-form text generation with citations.

\paragraph{Fact verification} Our approach is closely related to previous work on fact verification \citep{thorne-etal-2018-fever,wadden-etal-2020-fact,min-etal-2023-factscore}. One line of work focuses on evaluating outputs holistically with respect to the factuality criterion \cite{zhong-etal-2022-towards,liu-etal-2023-g,kim2024prometheus,kartac-etal-2025-openlgauge}. Fact verification is often also addressed using the decompose-then-verify approach, where the output is first decomposed into individual claims, followed by a verification of each claim against source texts \cite{fan-etal-2020-generating,wright-etal-2022-generating,min-etal-2023-factscore}.

Existing fact verification methods such as FactScore \citep{min-etal-2023-factscore} or SAFE \citep{wei2024long} treat each claim equally, which is not suitable for the medical domain. In contrast, VeriScore \citep{song-etal-2024-veriscore} limits claim extraction to only verifiable claims, while CORE \citep{jiang-etal-2025-core} filters claims based on their uniqueness and informativeness. However, this is still not suitable for our domain, where unverifiable, uninformative, or subjective claims are relatively rare, and hard filtering comes with a risk of false negatives, which should be avoided in high-stakes scenarios. Moreover, \citet{huang2025medscoregeneralizablefactualityevaluation} find that FactScore produces many invalid claims in the medical domain, including redundant, incomplete, or hallucinated claims, while both VeriScore and CORE suffer from frequent omissions. \citet{wanner2025claimsequalclaimsequal} classify claims based on their relevance to the user's query. However, our domain requires both risk-informed clinical importance (independent of the user's query) and fine-grained assessments.

\paragraph{NLP for veterinary medicine} Although substantially underrepresented compared to human medicine, NLP applications for veterinary medicine have received increased attention in recent years. Early work focused on data extraction \cite{anholt2014using,welsh2017disease} and classification \cite{dorea2013exploratory,anholt2014mining,awaysheh2016evaluation}, motivated primarily by epidemiology and surveillance. \citet{nie2018deeptag} applied a bidirectional LSTM model to automated diagnosis coding (prediction of standardized codes), while \citet{zhang2019vettag} used a Transformer model for the same problem. VetBERT \citep{hur2020domain} and PetBERT \citep{farrell2023petbert} are domain-adapted BERT models for disease syndrome classification and syndromic disease coding, respectively. More recent work has explored the use of LLMs for veterinary diagnostics \cite{abani2023chatgpt,okur2026comparison}, clinical information extraction \cite{fins2024evaluating,hur2024right,wulcan2025classification}, or diagnosis coding \cite{jiang2023vetllm,boguslav2026fine}. 
Existing evaluation studies are mostly limited to multiple-choice exams \citep{coleman2024two,alonso2025performance}.
\citet{farrell-etal-2025-peteval} introduced a benchmark derived from veterinary electronic health records, and \citet{kim-etal-2026-petqa} constructed the first veterinary benchmark focused on long-form question answering.

\section{Methods}
\label{sec:methods}

\subsection{Problem Statement}

Given a query $Q$ and a model $\mathcal{M}$, let $Y = \mathcal{M}(Q)$ be an answer to the query, consisting of tuples $\{(T_1, R_1), ..., (T_{|Y|}, R_{|Y|})\}$, where $T_i$ is the $i$-th text segment of $Y$ and $R_i = \{r_1, ..., r_M\}$ is the corresponding (possibly empty) set of $M$ citations with excerpts. The task is to provide a score that represents the degree to which $T_i$ is supported by $R_i$ in a way that weighs the information included in $T_i$ by its harm potential. In other words, we assess what fraction of the total harm potential in segment $T_i$ is supported. Information with higher harm potential should contribute a relatively larger penalty in the calculation of the verification score. Note that unlike fact-checking approaches, we focus on faithfulness: the verification is based only on the provided inline excerpts, although our method can be naturally extended to fact-checking based on external sources retrieved post hoc.

\subsection{\textsc{VetScore}}
\label{sec:methods_metric}

Our approach follows the decompose-then-verify paradigm \cite{min-etal-2023-factscore,wei2024long,song-etal-2024-veriscore}. We first decompose segment $T_i$ into a set of $N$ individual claims $C_i = \{c_1, ..., c_N\}$ and then use a verification function $\mathcal{V}$ to verify each claim $c_j$ against its cited references $R_i$ to obtain the verification score $v_j = \mathcal{V}(R_i, c_j) \in \{0, 1\}$, where 1 indicates that $c_j$ is supported by $R_i$, while 0 indicates that $c_j$ is not supported by, or contradicts $R_i$. Additionally, we apply weighting to each claim, which allows different contributions of claims to the final score based on their harm potential. We define $w_j = \mathcal{H}(c_j) \in \{1, \ldots, 5\}$ as a harm potential score for claim $c_j$, where $\mathcal{H}$ is a scoring function. Harm potential is assessed regardless of the actual correctness of the claim: the rater considers the extent of potential harm caused if the claim was wrong and a practitioner used it to inform their decisions. Applying $\mathcal{V}$ and $\mathcal{H}$ independently to each $c_j \in C_i$, we obtain a set of tuples $\{(v_{c_1}, w_{c_1}), ..., (v_{c_N}, w_{c_N})\}$. Finally, the combined scoring function $\mathcal{S}$ calculates the final risk-adjusted verification rate $\mathcal{S}(\{(v_{c_1}, w_{c_1}), ..., (v_{c_N}, w_{c_N})\})$. In the following, we describe all components in detail.

\paragraph{Output segmentation} We apply a regex to split the output text into segments with the corresponding citations and excerpts. We define a segment as a substring of the output between two delimiters, where a delimiter is an empty line or a previous citation. The immediately following citations are then associated with the segment. Markdown headers and purely formatting segments (such as horizontal lines) are excluded. Most resulting segments correspond approximately to a single sentence, but can be longer in case of multiple adjacent sentences without citations.

\paragraph{Claim decomposition} Given a text segment, we instruct an LLM to decompose it into claims. \citet{gunjal-durrett-2024-molecular} argue that atomic claims are not the correct representation for fact verification, as they tend to be overly decomposed and not properly decontextualized.
Therefore, we give the LLM additional context useful for resolving anaphora and other forms of contextual dependence, i.e., up to the last 2,000 characters or the last header. See Figure~\ref{fig:prompt_template_decomp} in the Appendix for the claim decomposition prompt template.

\paragraph{Fact verification} To obtain the binary verification score for a segment $T_i$, we apply an LLM to determine whether each claim $c_j \in C_i$ is supported by the corresponding references. The LLM is given a set of references and a set of individual claims derived from a specific segment and is instructed to decide for each claim whether it is supported by the given set of references. All claims for a specific segment are scored in a single inference call, where the LLM is instructed to output a JSON array with a separate object for each claim, which includes fields with the verbatim copy of the claim, short explanation, and the score. The prompt template is shown in Figure~\ref{fig:prompt_template_verification} in the Appendix. If $R_i = \varnothing$, the LLM inference is skipped and the score is set to $0$ for the given segment.

\paragraph{Harm potential scoring} The scoring function is implemented by prompting an LLM to assign a harm potential score $w_j$ on a 5-point scale to each claim $c_j$, where 1 corresponds to ``no clinical consequence'', while 5 indicates ``direct patient harm''. Similarly to fact verification, the LLM is instructed to provide a JSON array with copies of the claims, explanations, and scores. The prompt template is shown in Figure~\ref{fig:prompt_template_harm_potential} in the Appendix.

\paragraph{Combined score} Given a segment $T_i$ and the corresponding set of tuples with binary verification scores and ordinal harm potential scores $\{(v_{c_1}, w_{c_1}), ..., (v_{c_N}, w_{c_N})\}$, we define the \textit{absolute combined score} as follows:

\begin{equation}
\label{eq_4}
S_{abs}(T_i) = \sum_{j=1}^{|C_i|} w_{j}^{\,p} (1 - v_{j})
\end{equation}

\noindent This can be interpreted as the magnitude of the unverified harm potential in $T_i$. The exponent $p$ allows us to control the contribution of higher scores to the penalty relative to lower scores. We use $p = 2.426$ in all experiments.\footnote{The rationale behind this value is as follows. If all harm potential levels are equally represented within a subset of claims, we want the score of 5 to correspond to 50\% of the overall weight.} 

Since the absolute score is theoretically unbounded from above, we add a normalization which allows the score to be interpreted as the fraction of total harm potential in segment $T_i$ that is not supported, while ensuring that it always falls within $[0, 1]$. We first compute the \textit{local} penalty of the output segment $T_i$ as follows:

\begin{equation}
\label{eq_1}
p_{local}(T_i) = \sum_{j=1}^{|C_i|} \frac{w_j^{\,p} (1 - v_j)}{\sum_k w_k^{\,p}} 
\end{equation}

\noindent We also define a \textit{global} variant of the penalty, which is calculated over any $Y' \subseteq Y$:

\begin{equation}
\label{eq_2}
p_{global}(Y') = \sum_{i=1}^{|Y'|}\sum_{j=1}^{|C_{i}|} \frac{w_{ij}^{\,p} (1 - v_{ij})}{\sum_{k,l} w_{kl}^{\,p}}  
\end{equation}

\noindent If $|Y'| = 1$, this is equivalent to Equation~\ref{eq_1}.

The final risk-adjusted verification score for a subset $Y'$ is then calculated as:
\begin{equation}
\label{eq_3}
\mathcal{S}(Y') = 1 - p_{global}(Y')
\end{equation}

\section{Meta-evaluation dataset}
\label{sec:dataset}

To construct a meta-evaluation dataset to validate \textsc{VetScore}, we collect a set of input queries, retrieve sources relevant to these queries, and apply six different LLMs to generate outputs with citations based on the queries and the sources (Section~\ref{sec:output_generation}). We then conduct a human annotation with veterinary experts and students (Section~\ref{sec:human_annotation}).

\subsection{Output generation}
\label{sec:output_generation}

\paragraph{Input queries} The queries selected as our inputs are based on real production data from a veterinary QA system.\footnote{\url{https://www.primveterinary.com}} We first search by keywords to obtain queries related to latest evidence, recent guidelines, and treatment plans. This is followed by sampling 100 queries, filtering and paraphrasing. We ensure that queries are diverse, relevant, and representative, and do not contain sensitive information. The final subset after filtering consists of 67 unique queries that span various clinical disciplines (e.g., gastroenterology, cardiology, pharmacology, etc.), question types (e.g., treatment selection, guideline recommendation, drug safety, etc.), and patient species. See Appendix~\ref{sec:queries} for the complete list of queries and their categories.

\paragraph{System outputs} For output generation, we implement a simple system consisting of a retrieval module and an LLM for generating answers. The retrieval module is based on Exa Search\footnote{\url{https://exa.ai}} and an LLM-based relevance scoring component used for an additional filtering, keeping only sources labeled as either directly or partially relevant (the prompt template with the instructions is shown in Figure~\ref{fig:prompt_template_relevance} in the Appendix). We restrict the search to articles in the PubMed Central (PMC) archive.\footnote{\url{https://pmc.ncbi.nlm.nih.gov/}}
The number of sources in the LLM context is limited to a maximum of ten articles. Given the sources retrieved for a query, the models are instructed to generate a response to the query, citing each claim in the output with one or more excerpts from the source. To allow detection of potentially hallucinated sources, each excerpt should be followed by a source ID with line indices (see Figure~\ref{fig:diagram} for an example).
The answer for each query is generated with each of the following LLMs: Claude Opus 4.6,\footnote{\url{https://www.anthropic.com/claude/opus}} Gemini 3.1 Pro,\footnote{\url{https://deepmind.google/models/gemini/pro}} GPT-5.4,\footnote{\url{https://openai.com/index/introducing-gpt-5-4}} DeepSeek V3.2 671B \citep{deepseekai2025deepseekv32pushingfrontieropen}, Qwen3.5 397B A17B \citep{qwen35}, and Mistral Small 4 119B.\footnote{\url{https://mistral.ai/news/mistral-small-4}} We obtain 402 output texts overall (6 models $\times$ 67 queries), with an average of 31.5 ($\pm$16.2) excerpts per output and an average excerpt length of 128 ($\pm$13) characters. See Appendix~\ref{sec:system_outputs} for details on the inference and output text statistics. The prompt templates used to generate the outputs are shown in Figures~\ref{fig:prompt_template_output_system}--\ref{fig:prompt_template_output_user} in the Appendix.

\paragraph{Citation validity} For each citation and the corresponding excerpt, it only makes sense to consider segments with valid citations, while others should receive the lowest possible score. Thus, we use several metrics to identify invalid citations and excerpts: (1) \emph{No source}, the proportion of citations with invalid source ID, (2) \emph{Exact}, the proportion of excerpts matched verbatim in the corresponding source, and (3) \emph{Fuzzy}, the proportion of excerpts that could be matched if ellipses of the form \texttt{"..."} are allowed. For the \emph{Exact} and \emph{Fuzzy} metrics, we additionally measure the proportion of cases where the match is on any line in the source. The LLMs vary substantially with respect to the number of valid citations, from 93\% (Gemini 3.1 Pro) to only 42\% exact matches (Mistral Small 4). See Appendix~\ref{sec:citation_validity_details} for more details.

To prepare the final dataset, we segmented the generated outputs (see Section~\ref{sec:methods_metric}), sampled three random segments from each, and filtered out segments with invalid citations (i.e., those without at least a fuzzy match). Finally, we applied claim decomposition using Gemini 3 Flash to obtain individual claims. In total, the dataset includes 1,022 segments with 4,068 claims.

\subsection{Human annotation}
\label{sec:human_annotation}

We employed veterinary experts as annotators for two separate tasks: fact verification and harm potential scoring. The experts consisted of two groups: (1) veterinary students and recent graduates, and (2) professional practitioners with at least three years of experience. To reduce potential bias, each claim was annotated by three annotators (incl.\ at least one practitioner).
Before the main annotation, we conducted a pilot which served as a qualification task as well as a guideline refinement and ambiguity resolution stage. The annotators labeled the data in a web user interface in batches of 10 segments, and were shown one set of claims at a time. See Appendices~\ref{sec:annotator_details}--\ref{sec:annotation_setup} for details on the annotators, annotation interface, and annotation guidelines.
We use Krippendorff's $\alpha$ to assess the overall inter-annotator agreement (IAA) on the dataset.\footnote{Unlike Fleiss' $\kappa$, Krippendorff's $\alpha$ is more appropriate for measuring inter-annotator agreement with ordinal ratings \citep{hayes2007answering,artstein-poesio-2008-survey}.}

\paragraph{Task 1: Fact verification} In this task, each annotator was given a set of examples, where each example consisted of source excerpts (including source titles), input query,\footnote{We translated each query to a topic formulation, as we find that including queries directly in the user interface makes the task more difficult for the annotators.} and claims. The responses were binary (yes/no) and the annotators were instructed to decide for each claim whether it is supported by (or can be reasonably inferred from) the given source excerpts. The guidelines also included an instruction to rate the faithfulness of a claim to the provided excerpts regardless of the claim's actual correctness. We obtained $\alpha = 0.623$ on the verification task.

\paragraph{Task 2: Harm potential scoring} To annotate harm potential, annotators were given an input query and claims derived from the original segment and were asked to rate each claim on a scale between 1 (no clinical consequence) and 5 (direct patient harm). Specifically, the task was to decide how much harm could result if the claim was wrong and a veterinary practitioner using the system acted on it (see Section~\ref{sec:methods_metric}). The annotators were instructed to rate each claim regardless of its correctness, and unlike in Task 1, they were not shown the associated source excerpts as these are not relevant for the subtask. We obtained the agreement of $\alpha = 0.467$ on the scoring task.

\subsection{Multi-facet Rasch model}
\label{sec:item_response_theory}

We apply item response theory (IRT, \citealp{rasch1968mathematical,de2009theory,lalor2016buildingevaluationscaleusing}), to analyze possible causes of lower IAA and to aggregate individual annotation scores (see Section~\ref{sec:score_aggregation}). Specifically, we use the partial-credit multi-facet Rasch model (MFRM, \citealp{linacre1994mfrm}), a type of polytomous Rasch model \citep{andrich1978rating}, which allows us to estimate the latent harm potential of each claim and severity of each annotator (interpreted as sensitivity to potential harm) on a common logit scale. Moreover, it estimates the effective thresholds, i.e., locations on the scale above which a given annotator is more likely to assign the claim a score $k$ as opposed to $k - 1$. MFRM estimates the probability that the annotator $a$ with thresholds $\tau_{a,1}, ..., \tau_{a,k}$ assigns a score $k$ to claim $n$ with latent harm potential $\theta_n$:
\begin{equation}
\label{eq_5}
    P_a(k | \theta_n) \propto \exp\left( k \cdot \theta_n -  \sum\nolimits_{j \leq k} \tau_{a,j} \right)
\end{equation}

\begin{figure}[t]
  \includegraphics[width=\columnwidth]{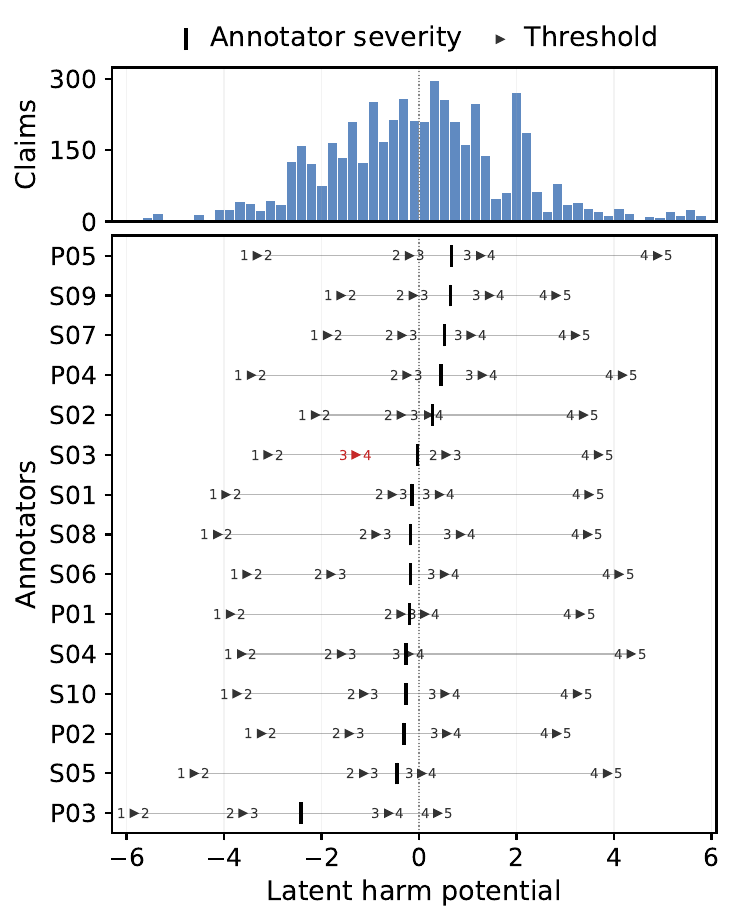}
  \caption{Wright map for the MFRM model fitted on our annotations of the harm potential scoring task. \textit{Top:} Latent harm potential distribution of annotated claims. \textit{Bottom:} Annotator severity and effective thresholds. Annotators are labeled as students (S) and practitioners (P). Red arrow indicates a disordered threshold.}
  \label{fig:wright_map}
\end{figure}

We use the TAM package\footnote{\url{https://cran.r-project.org/package=TAM}} and estimate the model as a partial credit model (PCM,  \citealp{masters1982rasch}) with a raters-as-items parametrization, which is equivalent to MFRM with one task/item and rater-specific thresholds \citep{robitzsch2018item}. Figure~\ref{fig:wright_map} shows a Wright map \citep{wilson2023constructing} for the model. Although the thresholds of the individual annotators differ substantially, they are well-ordered (with a single exception of S03), and their sensitivities are close to each other. Annotator P03 is an outlier, shown by lower pairwise agreements with several other annotators (see Figure~\ref{fig:expert_agreements} in the Appendix). Appendix~\ref{sec:annotation_aggregation} presents a detailed analysis and additional details on the method. Our results suggest that the observed differences are systematic and reflect a different use of the scale by individual annotators rather than inconsistent scoring.

\subsection{Annotator score aggregation}
\label{sec:score_aggregation}

To aggregate the scores for fact verification, we average the binary labels to obtain a continuous score between 0 and 1. For the harm potential scoring task, we use the MFRM model described above to obtain canonical rater thresholds $\bar{\tau}_k = 1/\mathcal{A}\sum_{a \in \mathcal{A}} \tau_{a,k}$, where $\mathcal{A}$ is the set of all annotators, and then use Equation~\ref{eq_5} to calculate the aggregate score for claim $n$ as $\sum_k k \cdot P(k | \theta_n)$.

\subsection{Evaluation of claim decomposition}
\label{sec:claim_decomp_eval}

Since the effectiveness of verification strongly depends on the decomposition method \citep{wanner-etal-2024-closer}, we manually validate the quality of the decomposition step. We examined 120 randomly selected segments and their decompositions. Although we found 11 segments (9\%) with issues, these are mostly minor. We identify three categories of issues: (1) \emph{semantic drifts} (5 examples) that slightly change the meaning of the original claim, (2) \emph{suboptimal decomposition} (3 examples), where the original segment is not fully decomposed, and (3) \emph{contextualization issues} (2 examples) where an anaphora is not resolved. Additionally, we found one case of uninformative claims. No hallucinations were found in the analyzed decompositions. Further analysis shows that these issues only have a small effect on the combined score (see Appendix~\ref{sec:decomposition_analysis}).

\begin{table*}[t]
    \centering
    \small
    \setlength{\tabcolsep}{6pt}
    \begin{tabular}{lccccc}
    \toprule
    \textbf{Model} & \shortstack{\textbf{Fact}\\ \textbf{verification}} & \shortstack{\textbf{Harm}\\ \textbf{potential}} & \shortstack{\textbf{Combined}\\\textbf{(local)}} & \shortstack{\textbf{Combined}\\\textbf{(global)}} & \shortstack{\textbf{Combined}\\\textbf{(absolute)}} \\
    \midrule
    \multicolumn{6}{l}{\textsc{VetScore}} \\
    \noalign{\vskip 1pt}
    GPT-5.6 Sol        & $0.758_{\pm 0.004}$ & $0.713_{\pm 0.003}$ & $0.738_{\pm 0.003}$ & $0.779_{\pm 0.006}$ & $0.732_{\pm 0.005}$ \\
    Gemini 3.1 Pro (low) & $0.781_{\pm 0.004}$ & $0.761_{\pm 0.003}$ & $0.757_{\pm 0.006}$ & $0.778_{\pm 0.006}$ & $0.707_{\pm 0.004}$ \\
    Gemini 3.6 Flash   & $0.800_{\pm 0.002}$ & $0.732_{\pm 0.003}$ & $0.776_{\pm 0.004}$ & $0.798_{\pm 0.005}$ & $0.750_{\pm 0.005}$ \\
    \noalign{\vskip 2pt}
    \hdashline[0.5pt/1.5pt]
    \noalign{\vskip 3pt}
    Gemma 4 31B        & $0.739_{\pm 0.002}$ & $0.719_{\pm 0.003}$ & $0.716_{\pm 0.003}$ & $0.731_{\pm 0.006}$ & $0.684_{\pm 0.003}$ \\
    Qwen3.5 35B A3B    & $0.702_{\pm 0.008}$ & $0.677_{\pm 0.004}$ & $0.678_{\pm 0.014}$ & $0.698_{\pm 0.020}$ & $0.648_{\pm 0.010}$ \\
    \noalign{\vskip 2pt}
    \hdashline[0.5pt/1.5pt]
    \noalign{\vskip 3pt}
    Baichuan-M3 235B   & $0.561_{\pm 0.011}$ & $0.678_{\pm 0.011}$ & $0.547_{\pm 0.012}$ & $0.592_{\pm 0.024}$ & $0.559_{\pm 0.019}$ \\
    MedGemma 27B       & $0.446_{\pm 0.009}$ & $0.660_{\pm 0.008}$ & $0.415_{\pm 0.006}$ & $0.444_{\pm 0.027}$ & $0.395_{\pm 0.015}$ \\
    \midrule
    \multicolumn{6}{l}{\textit{Verification-only}} \\
    \noalign{\vskip 1pt}
    GPT-5.6 Sol        & $0.758_{\pm 0.004}$ & - & $0.743_{\pm 0.001}$ & $0.776_{\pm 0.004}$ & $0.669^{***}_{\pm 0.004}$ \\
    Gemini 3.6 Flash   & $0.800_{\pm 0.002}$ & - & $0.787_{\pm 0.003}$ & $0.814_{\pm 0.005}$ & $0.683^{***}_{\pm 0.007}$ \\
    \noalign{\vskip 2pt}
    \hdashline[0.5pt/1.5pt]
    \noalign{\vskip 3pt}
    Gemma 4 31B        & $0.739_{\pm 0.002}$ & - & $0.725_{\pm 0.002}$ & $0.742_{\pm 0.006}$ & $0.632^{***}_{\pm 0.001}$ \\
    Qwen3.5 35B A3B    & $0.702_{\pm 0.008}$ & - & $0.682_{\pm 0.011}$ & $0.713_{\pm 0.016}$ & $0.588^{***}_{\pm 0.010}$ \\
    \midrule
    \multicolumn{6}{l}{\textit{End-to-end}} \\
    \noalign{\vskip 2pt}
    GPT-5.6 Sol        & - & - & $0.319^{***}_{\pm 0.004}$ & $0.289^{***}_{\pm 0.013}$ & - \\
    Gemini 3.6 Flash   & - & - & $0.452^{***}_{\pm 0.012}$ & $0.454^{***}_{\pm 0.008}$ & - \\
    \noalign{\vskip 2pt}
    \hdashline[0.5pt/1.5pt]
    \noalign{\vskip 3pt}
    Gemma 4 31B        & - & - & $0.485^{***}_{\pm 0.014}$ & $0.464^{***}_{\pm 0.013}$ & - \\
    Qwen3.5 35B A3B    & - & - & $0.313^{***}_{\pm 0.015}$ & $0.305^{***}_{\pm 0.036}$ & - \\
    \bottomrule
    \end{tabular}
    \caption{Spearman ($\rho$) correlations with human annotations for the full \textsc{VetScore} pipeline and two baselines. \textit{Verification-only} drops the harm-potential component and scores each segment by fact verification alone. \textit{End-to-end} produces the combined score in a single call, without decomposition. Each \textit{Combined} column correlates the setup's score with the human combined score, normalized \textit{locally} (per segment) or \textit{globally} (per output), or in the non-normalized \textit{absolute} form. Each cell is the mean over $n=5$ runs with the standard deviation as a subscript. In the lower two sections, asterisks mark a significant difference of the \textit{Combined} score from the same model's \textsc{VetScore} value (paired bootstrap over segments, 10,000 resamples: $^{*}:p<0.05$, $^{**}:p<0.01$, $^{***}:p<0.001$). The table presents a subset of judge models, see Appendix~\ref{sec:full_results} for the complete set.}
    \label{tab:main_results}
    \end{table*}

\section{Experiments}
\label{sec:experiments}

\paragraph{Baselines} We compare \textsc{VetScore} with the following baselines. In the \emph{verification-only} setup, we exclude the harm potential scoring component. In the \textit{end-to-end} baseline (see Appendix~\ref{sec:prompt_templates}), the LLM is prompted to provide the overall risk-adjusted verification score in range [0, 1].

\paragraph{Judge models} Since the outputs for veterinary long-form QA can be relatively long and our setup analyzes these outputs at the segment level, we focus on smaller LLMs to keep our approach efficient. We use claims as decomposed by Gemini 3 Flash (since these were used to create our annotation dataset), and run verification and harm potential scoring with each of the following LLMs as a judge model: proprietary models GPT-5.6 Sol,\footnote{\url{https://openai.com/index/gpt-5-6/}} Claude Sonnet 5\footnote{\url{https://www.anthropic.com/claude/sonnet}} and Gemini 3.1 Pro\footnote{\url{https://deepmind.google/models/gemini/pro/}} to estimate the current performance upper bound for our approach; Gemini 3 Flash and Gemini 3.6 Flash\footnote{\url{https://deepmind.google/models/gemini/flash/}} as more efficient alternatives; Gemma 4 31B \citep{gemmateam2026gemma4technicalreport}, Qwen3.5 35B A3B \citep{qwen35}, Nemotron 3 Nano 30B A3B \citep{nvidia2025nemotron3nanoopen}, and GLM-4.7 Flash \citep{Zeng2025GLM45AR}, which represent smaller open-weight models; and Baichuan-M3 235B \citep{m3team2026baichuanm3modelingclinicalinquiry}, MedGemma 27B, and MedGemma 1.5 4B \citep{sellergren2026medgemmatechnicalreport} as specialized models fine-tuned for medical tasks. We score only segments containing valid citations. See Appendix~\ref{sec:model_inference} for details on inference.

\paragraph{Meta-evaluation} We evaluate the individual components of \textsc{VetScore} through claim-level Spearman correlation of the metric scores with human expert scores. We also evaluate combined scores obtained through aggregation (see Section~\ref{sec:methods_metric}) by measuring Spearman correlation with combined scores derived from human annotation.
As each claim is annotated three times, we aggregate human annotations for each claim, both for fact verification and for harm potential scoring, following Section~\ref{sec:score_aggregation}. Finally, we measure precision and recall on the identification of high-risk segments, which we define as segments with at least one unsupported claim with harm potential score larger than or equal to a threshold $t$. We use $t = 4$ for judges and $t = 3.43$  for human scores.\footnote{Our MFRM model (see Section~\ref{sec:item_response_theory}) gives $P(\text{score} \geq 4) = 0.5$ with an aggregate score value of 3.43.}

\section{Results}
\label{sec:results}

\paragraph{Main results} Table~\ref{tab:main_results} shows the Spearman correlations between human experts and \textsc{VetScore} with different LLMs used for fact verification and harm potential scoring. The last three columns show the correlations between the \textit{local}, \textit{global},\footnote{Note that we sample three segments from each output, therefore \emph{global} combined scores are calculated over these subsets of segments.} and \textit{absolute} combined scores. We run all models with sampling ($n = 5$), measure the correlation for each $n$ separately, and then average the correlations. Across most judge models, our approach achieves good correlation with humans: up to 0.8 for fact verification, 0.76 for harm potential scoring, and 0.8 for the combined score, while the variance across runs typically stays low. Gemini 3.6 Flash achieves the highest correlation in fact verification and the combined score, while Gemini 3.1 Pro surpasses all judge models on harm potential scoring. The open-weight Gemma 4 31B is competitive with proprietary models despite its relatively small parameter size, which suggests that our modular approach can make the evaluation work reasonably well even with smaller and more efficient LLMs (although the results are considerably worse for two MoE models -- GLM-4.7 Flash and Nemotron 3 Nano -- see Table~\ref{tab:full_results} in the Appendix). Specialized medical LLMs perform considerably worse as judges on fact verification, although Baichuan-M3 and MedGemma 27B show reasonable correlations with humans on harm potential scores.\footnote{We use a 4-bit quantized version of Baichuan-M3 235B (see Appendix~\ref{sec:model_inference}) and both MedGemma models are based on an older version of Gemma. These factors could have contributed to the worse results observed with the specialized medical models.} The \textit{verification-only} baseline shows substantially lower correlations in combined absolute scores compared to our main setup. Note that the effect is not present in correlations with weighted scores, where the harm magnitude is not preserved, which allows the verification scores to dominate the correlation results. The \textit{end-to-end} baseline shows relatively low correlations with human experts, which is expected given the task complexity.

\begin{figure}[t]
  \includegraphics[width=\columnwidth]{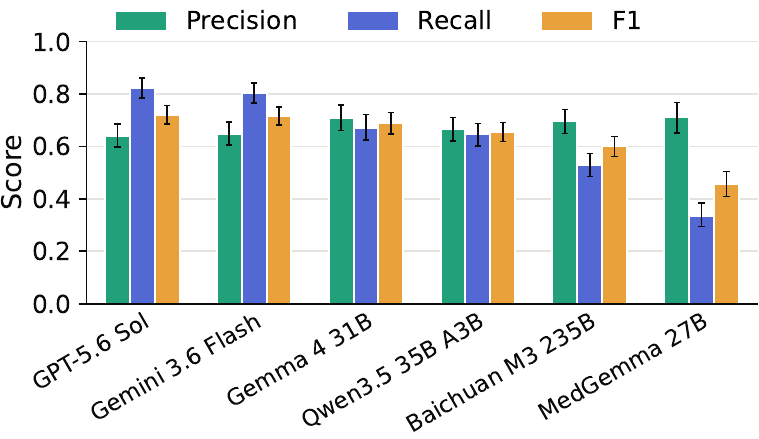}
  \caption{Precision, recall, and F1 scores for a subset of judge models on high-risk segment identification, where a text segment is flagged if it contains any unsupported claim with a harm potential score $\geq t$, with $t = 4 $ for judges and $t = 3.43$ for humans (see Section~\ref{sec:experiments}). Error bars indicate 95\% bootstrap CIs (1,000 resamples).}
  \label{fig:risk_detection}
\end{figure}

Figure~\ref{fig:risk_detection} shows results on identification of high-risk segments. With the strongest judge models, our approach achieves good recall ($\geq 0.8$), while the precision is slightly lower and stays similar across models. Appendix~\ref{sec:risk_identification} presents results for all judge models with additional analysis, showing that LLMs are generally stricter in their harm potential scoring compared to humans, which explains the lower precision for the strongest judge models (see Table~\ref{tab:risk_identification}). In addition, we compare \textsc{VetScore} with a verification-only baseline, which leads to noisier identification with considerably lower precision (see Figure~\ref{fig:risk_detection_verify_only} in the Appendix).

Table~\ref{tab:system_rankings} shows how each system is scored and ranked by each judge. The results indicate a lack of self-preference bias \citep{panickssery2024llm} within model families: judges do not rank outputs of models from their own family significantly higher than other judges, which is likely a consequence of the decomposition to claims.

\paragraph{Ablations} To understand how different components of our pipeline contribute to its correlation with humans, we run a set of ablation experiments with two LLMs (see Table~\ref{tab:ablation_results}). In \textit{No reasoning field}, we remove the \texttt{reasoning} field from the output schema. The effect depends on the judge model, decreasing the fact verification correlation only for Gemini 3.6 Flash, and harm potential correlation only for Gemma 4 31B. Although the overall effect is small, the field is important for the explainability of our approach and produces only a small number of additional tokens. \textit{Joint evaluation} combines the instructions for the verification and scoring tasks into a single prompt and inference run, which has a significant effect on fact verification for Gemini 3.6 Flash. \textit{Simple rubric} removes the detailed rubric from the harm potential scoring prompt, substantially decreasing the correlation for both models.

\begin{table}[t]
\centering
\small
\setlength{\tabcolsep}{4pt}
\begin{tabular}{l>{\hspace{3mm}}l>{\hspace{2mm}}l>{\hspace{3mm}}l}
\toprule
\textbf{Model} & \hspace{-3mm}\shortstack{\textbf{Fact}\\ \textbf{verification}} & \hspace{-2mm}\shortstack{\textbf{Harm}\\ \textbf{potential}} & \hspace{-3mm}\shortstack{\textbf{Combined}\\\textbf{(absolute)}} \\
\midrule
\multicolumn{4}{l}{\textsc{VetScore}} \\
Gemini 3.6 Flash   & $0.800$ & $0.732$ & $0.750$ \\
Gemma 4 31B        & $0.739$ & $0.719$ & $0.684$ \\
\midrule
\multicolumn{4}{l}{\textit{No reasoning field}} \\
Gemini 3.6 Flash   & $0.783^{**}$ & $0.732$ & $0.730^{*}$ \\
Gemma 4 31B        & $0.742$ & $0.702^{*}$ & $0.680$ \\
\midrule
\multicolumn{4}{l}{\textit{Joint evaluation}} \\
Gemini 3.6 Flash   & $0.774^{***}$ & $0.718$ & $0.725^{*}$ \\
Gemma 4 31B        & $0.728$ & $0.718$ & $0.708$ \\
\midrule
\multicolumn{4}{l}{\textit{Simple rubric}} \\
Gemini 3.6 Flash   & $0.800$ & $0.670^{***}$ & $0.743$ \\
Gemma 4 31B        & $0.739$ & $0.673^{***}$ & $0.682$ \\
\bottomrule
\end{tabular}
\caption{Spearman ($\rho$) correlations with human annotations for our ablation experiments. Values are means across $n=5$ runs. Asterisks indicate a significant difference from \textsc{VetScore} (paired bootstrap over segments, 10,000 resamples: $^{*}:p<0.05$, $^{**}:p<0.01$, $^{***}:p<0.001$).}
\label{tab:ablation_results}
\end{table}

\section{Analysis}
\label{sec:analysis}

\paragraph{Scoring distributions} We analyze the scoring distributions of human annotators and models and find that humans are generally stricter in assessing whether a claim is supported by an excerpt (Figure~\ref{fig:human_distributions}), while the worst-performing judge models are too lenient (Figure~\ref{fig:verification_distributions}). When scoring the harm potential, human experts focus on the middle portion of the scale (2-4), while LLMs have a mode at 4, underutilizing the highest score. See Appendix~\ref{sec:score_distributions} for details.

\paragraph{Rating consistency} We also analyze the consistency of different judge models across runs, and find that the outputs in both tasks are relatively inconsistent with most open-weight judge models, but become very reliable with Gemma 4 31B and all proprietary models (see Appendix~\ref{sec:consistency}). 

\paragraph{Sensitivity to weight exponent \emph{p}} To estimate the effect of \emph{p} used in the combined score (see Section~\ref{sec:methods_metric}), we compare the correlations of all judges with experts for each $p \in \{1, 2, 2.426, 3\}$. Table~\ref{tab:exponent_sensitivity} in the Appendix shows that the correlations and judge rankings differ only slightly between different values of \emph{p}.

\paragraph{Effect of models on claim decomposition} Since re-annotating the dataset with claims obtained by different models for the decomposition step would be prohibitively expensive, we were not able to compare different models on this step. As an approximation, we calculate correlations of combined scores from the expert annotation to those based on running the whole pipeline, including claim decomposition, with two other models: Gemini 3.6 Flash and Gemma 4 31B. We observe a difference of 0.04-0.05 in Spearman correlations on the absolute combined scores (see Table~\ref{tab:selfdecomp_results} in the Appendix). However, this analysis cannot reliably separate the effects of the decomposition quality from the disagreement with the set annotated by humans.

\section{Conclusion}
\label{sec:conclusion}

In this work, we present a method designed to evaluate faithfulness of generated answers in veterinary QA to their citation excerpts, providing risk-adjusted scores that weight each claim by its harm potential. We show that it achieves high correlation with human veterinary experts, achieves relatively good recall on high-risk claim identification, and is explainable on several levels: it decomposes outputs to individual claims, uses aspect-specific modules (verification and harm potential), and finally provides concise natural language explanation with only a minimal impact on efficiency. Although performance, reliability, and efficiency depend on the underlying judge model, our experiments show that the modularity of our approach allows its effective use with smaller models. In addition, we analyze the collected expert annotations using item response theory, which allows us to adjust for annotator-specific use of the harm potential scoring scale.

\section*{Limitations}

In this work, we focus on excerpt-based verification, which is a faithfulness problem, rather than traditional fact-checking. Although our preliminary experiments suggest that our approach could be easily extended to fact-checking against external sources by adding a retrieval module, we leave a rigorous evaluation of such an extension for future work. Also, since \textsc{VetScore} is focused on evaluating only claims that are present in the output text, it does not address omissions.

Our approach of risk-adjusted fact verification could in principle be generalized to human medicine. However, despite the close relation between these domains, there are substantial differences between veterinary and human medicine, and therefore, the method remains to be validated specifically for this domain.

We focus on text-only queries that involve latest evidence, guidelines, protocols, and general recommendations -- we leave complex scenarios, such as multi-turn interaction that includes patient history and multi-modal inputs, to future work.

While our item response theory analysis suggests that the differences observed between the annotators on the harm potential task are systematic and reflect different annotator sensitivity rather than inconsistent annotation, we consider a lower inter-annotator agreement ($\alpha = 0.467$) for this task a limitation of our work.

Although we do not annotate holistic segment-level scores due to complexity of such a task, it would be interesting to conduct an extrinsic evaluation of these combined scores as provided by \textsc{VetScore} to learn more about its usefulness for veterinary practitioners in verifying generated outputs with citations.

Our method provides concise natural language explanations generated by an LLM judge model. However, the faithfulness of these explanations to the provided scores has not been evaluated in this work.

Finally, while we include small and efficient judge models in our meta-evaluation, we have not explored training specialized models for the task (e.g. through distillation), which could make the setup even more efficient.

\section*{Ethical Considerations}

Since our work focuses on a high-stakes domain, we conduct a large-scale human annotation with veterinary experts (Section~\ref{sec:human_annotation}) to validate the performance of our approach. However, we use LLMs as underlying models, which are known to demonstrate various biases across many domains and tasks, including medicine. While we aim to mitigate this by surfacing the bias through a greater explainability compared to the conventional LLM-as-a-judge paradigm, the bias cannot be completely eliminated. For example, it cannot be guaranteed that the natural language explanations provided by the judge models are always faithful to their decisions. Additionally, our dataset reflects the frequencies of query types, disciplines, or species observed in real-world usage, making many of them underrepresented (Section~\ref{sec:output_generation}).

Although we include proprietary LLMs as judge models in our comparison, we focus on open-weight models to increase transparency and reproducibility.

We acknowledge the use of Claude Code in drafting parts of our code. Parts of the manuscript were revised using an AI-assisted grammar checker.

\section*{Acknowledgments}

We thank the veterinary expert annotators who contributed to this work. This research was partially funded by the European Union (ERC, NG-NLG, 101039303) and Charles University SVV project number 260 821. Expert annotation and all API costs were funded by PrimVeterinary. Experiments with local LLMs used resources of the LINDAT/CLARIAH-CZ Research Infrastructure (Czech Ministry of Education, Youth, and Sports project No. LM2023062).

\bibliography{custom}

\appendix

\section{Prompt Templates}
\label{sec:prompt_templates}

Prompt templates used for claim decomposition, fact verification and harm potential scoring used in \textsc{VetScore} are presented in Figures~\ref{fig:prompt_template_decomp}--\ref{fig:prompt_template_harm_potential}. Figures~\ref{fig:prompt_template_output_system} and \ref{fig:prompt_template_output_user} show the system and user prompts for generating system outputs for our dataset. The prompt used for the relevance assessment of retrieved sources is shown in Figure~\ref{fig:prompt_template_relevance}, and the prompt for the end-to-end baseline is shown in Figure~\ref{fig:prompt_template_end_to_end}. 

\section{Dataset details}
\label{sec:dataset_details}

\subsection{Input queries}
\label{sec:queries}

\begin{figure*}[t]
  \centering
  \includegraphics[width=\textwidth]{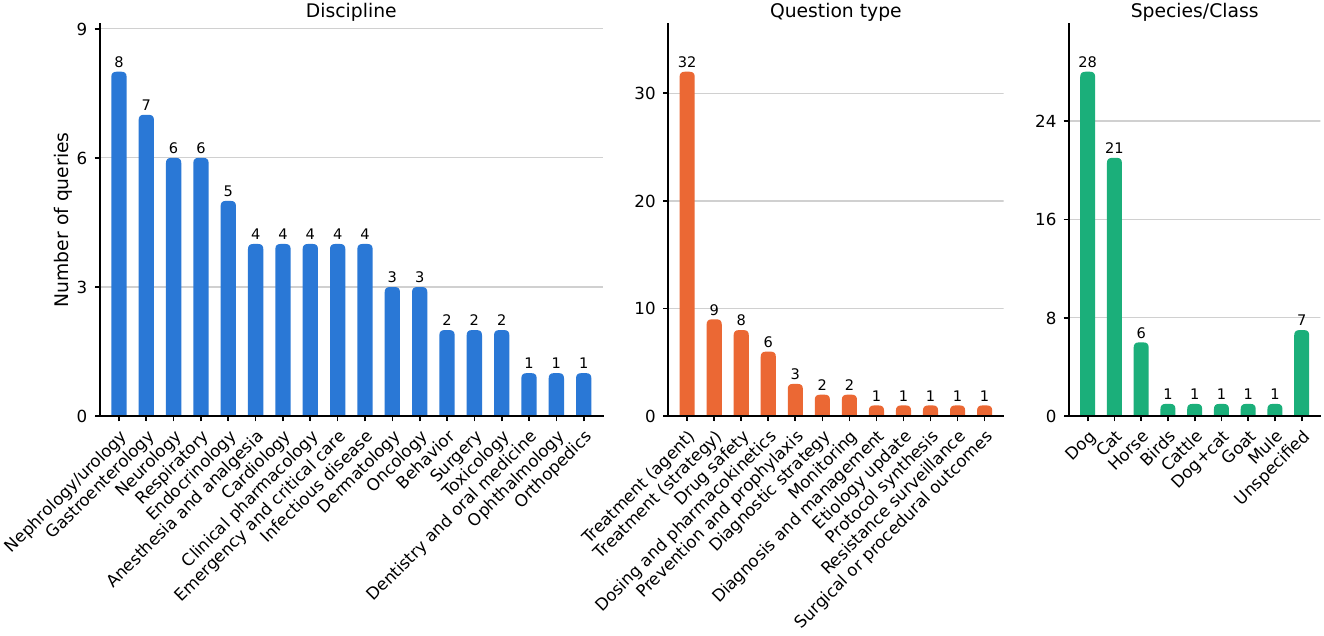}
  \caption{Distribution of disciplines, question types, and patient species represented in our input queries.}
  \label{fig:query_distributions}
\end{figure*}

Table~\ref{tab:queries} lists the queries we used as inputs to generate the outputs in our dataset. They are classified along three axes: discipline (gastroenterology, neurology, respiratory, etc.), question type (treatment, drug safety, diagnostic strategy, etc.), and species/class. Figure~\ref{fig:query_distributions} shows the distribution of these three categories. Since queries are sampled from a production veterinary QA system, the question type and species distributions are skewed, reflecting the real-world use of such systems. The queries were collected according to the PrimVeterinary Terms of Use.

\subsection{System outputs}
\label{sec:system_outputs}

We generate the system outputs with six different LLMs, using the OpenRouter\footnote{\url{https://openrouter.ai}} service. The LLMs are listed in Table~\ref{tab:output_models}, including the temperature and reasoning effort used for each model and its OpenRouter tag. For each LLM, we used the official recommended temperature. We set reasoning effort to \texttt{high} for all models that support this parameter.

\begin{figure*}[t]
  \centering
  \includegraphics[width=\columnwidth]{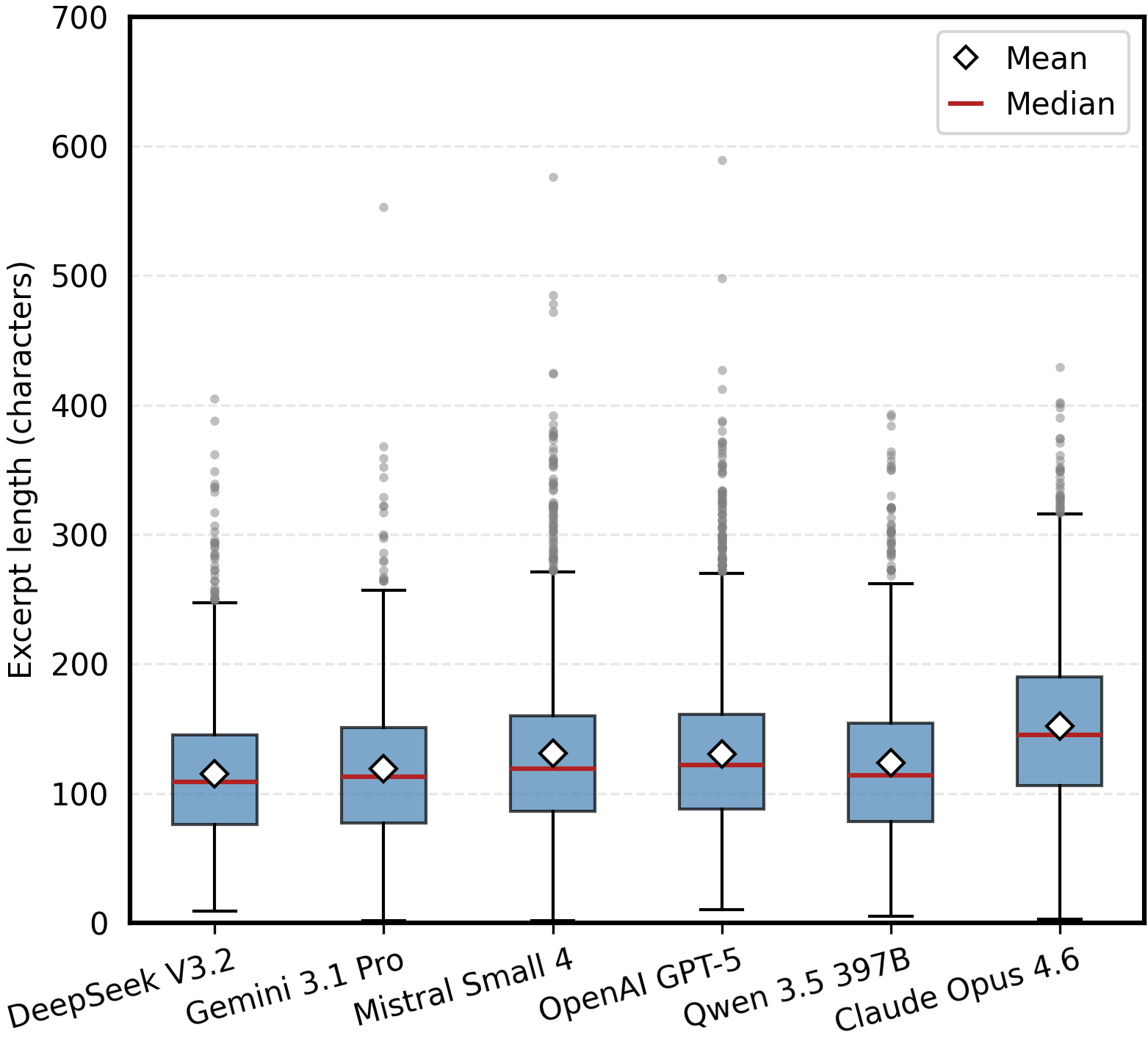}
  \caption{Distribution of citation excerpt lengths for the six LLMs across the 67 queries. Boxes show the inter-quartile range, whiskers extend to 1.5$\times$IQR, and dots are outliers. The $y$-axis is clipped at 700 characters, rare excerpts above that range are not shown.}
  \label{fig:excerpt_length}
\end{figure*}

Figure~\ref{fig:excerpt_length} shows the distribution of lengths of the citation excerpt in the outputs of the individual LLMs.

\subsection{Citation validity}
\label{sec:citation_validity_details}

\begin{table*}[h]
\centering
\small
\begin{tabular}{lcccccc}
\toprule
 & & \multicolumn{3}{c}{Regex} & \multicolumn{2}{c}{Semantic} \\
\cmidrule(lr){3-5} \cmidrule(lr){6-7}
Model & Citations & No source & Exact \% (+off) & Fuzzy \% (+off) & Fuzzy+format \% & Fuzzy+para \% \\
\midrule
Gemini 3.1 Pro    & 1106 & 0 (0.0\%) & \textbf{93.3 (94.1)} & \textbf{95.6 (96.5)} & \textbf{98.3} & \textbf{99.8} \\
Claude Opus 4.6   & 1983 & 0 (0.0\%) & 88.5 (89.6) & 92.0 (93.1) & 96.6 & 99.7 \\
Qwen 3.5 397B     & 2378 & 12 (0.5\%) & 84.4 (87.3) & 86.9 (90.0) & 93.8 & 98.3 \\
OpenAI GPT-5.4      & 3478 & 1 (0.0\%) & 80.2 (80.7) & 94.5 (95.2) & 96.5 & 99.3 \\
DeepSeek V3.2     & 1590 & 3 (0.2\%) & 71.4 (77.0) & 82.1 (89.4) & 91.0 & 98.1 \\
Mistral Small 4     & 2118 & 44 (2.1\%) & 41.5 (49.7) & 48.9 (59.5) & 62.2 & 85.5 \\
\bottomrule
\end{tabular}
\caption{Citation validity metrics for the generated dataset. The \textit{Regex} columns are computed by string matching against the cited source, while the \textit{Semantic} columns additionally apply an LLM classifier (Gemini 3.1 Pro) to unmatched cases. \textit{No source}: non-existent source ID. \textit{Exact \%}: excerpt found verbatim on the cited line(s). \textit{Exact+off \%}: excerpt found verbatim anywhere in the cited source. \textit{Fuzzy} variants additionally allow ellipsis-bridged anchors. \textit{Fuzzy+format \%}: \textit{Fuzzy (+off)} plus the non-matching excerpts that the LLM judge classifies as verbatim with only formatting differences. \textit{Fuzzy+para \%} further allows faithful paraphrases.}
\label{tab:citation_metrics}
\end{table*}

Table~\ref{tab:citation_metrics} shows the citation validity of each model that generated the system outputs. The LLMs vary substantially with respect to the number of valid citations: excerpts provided by Gemini 3.1 Pro have 93\% exact matches (97\% if we allow ellipses in the form \texttt{``\ldots''} and any line in the source), while only 42\% excerpts are matched exactly for Mistral Small 4. GPT-5.4 tends to use ellipses frequently, illustrated by a substantial gap between the prevalence of exact (80\%) and fuzzy (95\%) matches. Non-existent source IDs are rare, with the highest frequency (2\%) in the outputs of Mistral Small.

To further analyze the unmatched cases, we apply an LLM classifier (Gemini 3.1 Pro) to classify them into the following three categories: (1) \emph{formatting}, where the excerpt is found in the source with only formatting differences, (2) \emph{paraphrase}, where the excerpt is a faithful paraphrase of a part of the source, and (3) \emph{hallucination} where the excerpt or its part is fabricated. To validate the classifier, we manually annotated 100 randomly selected excerpts. The classifier achieves accuracy and a macro F1 score of 89\% (see Table~\ref{tab:classifier_eval}). As the results in Table~\ref{tab:citation_metrics} show, most unmatched excerpts contain either formatting differences or are faithful paraphrases of the source. Mistral Small is an exception: it paraphrases around 23\% excerpts, while more than 14\% of its excerpts are fabricated.

\begin{table}[t]
\centering
\small
\begin{tabular}{lcccc}
\toprule
\textbf{Class} & \textbf{Precision} & \textbf{Recall} & \textbf{F1} & \textbf{\#} \\
\midrule
Formatting    & 0.97 & 0.84 & 0.90 & 37 \\
Paraphrase    & 0.84 & 0.95 & 0.89 & 44 \\
Hallucination & 0.89 & 0.84 & 0.86 & 19 \\
\bottomrule
\end{tabular}
\caption{Evaluation results for the LLM classifier of~unmatched excerpts. The overall accuracy and the macro F1-score are both 0.89.}
\label{tab:classifier_eval}
\end{table}

\section{Annotator details}
\label{sec:annotator_details}

Our dataset was annotated by 18 expert annotators, of which 6 were professional practitioners with at least three years of experience, and 12 were students or recent graduates. All students were in the third or fourth year of their study of veterinary medicine. Annotators studied or practiced in one of the following countries: United States (7), Czech Republic (3), United Kingdom (3), Canada (2), Germany (1), Singapore (1), or Australia (1).

Before collecting our final dataset, we conducted a pilot annotation round with 10 segments (35 claims) for the verification task and 20 segments (107 claims) for the harm potential scoring task. We consulted on the guidelines and the annotation process with three veterinary practitioners and two students. Based on this, we obtained labels for the pilot subset and used the pilot as a qualification for additional annotators. None of the pilot segments were used in the main annotation.

\begin{figure*}[t]
  \centering
  \includegraphics[width=\textwidth]{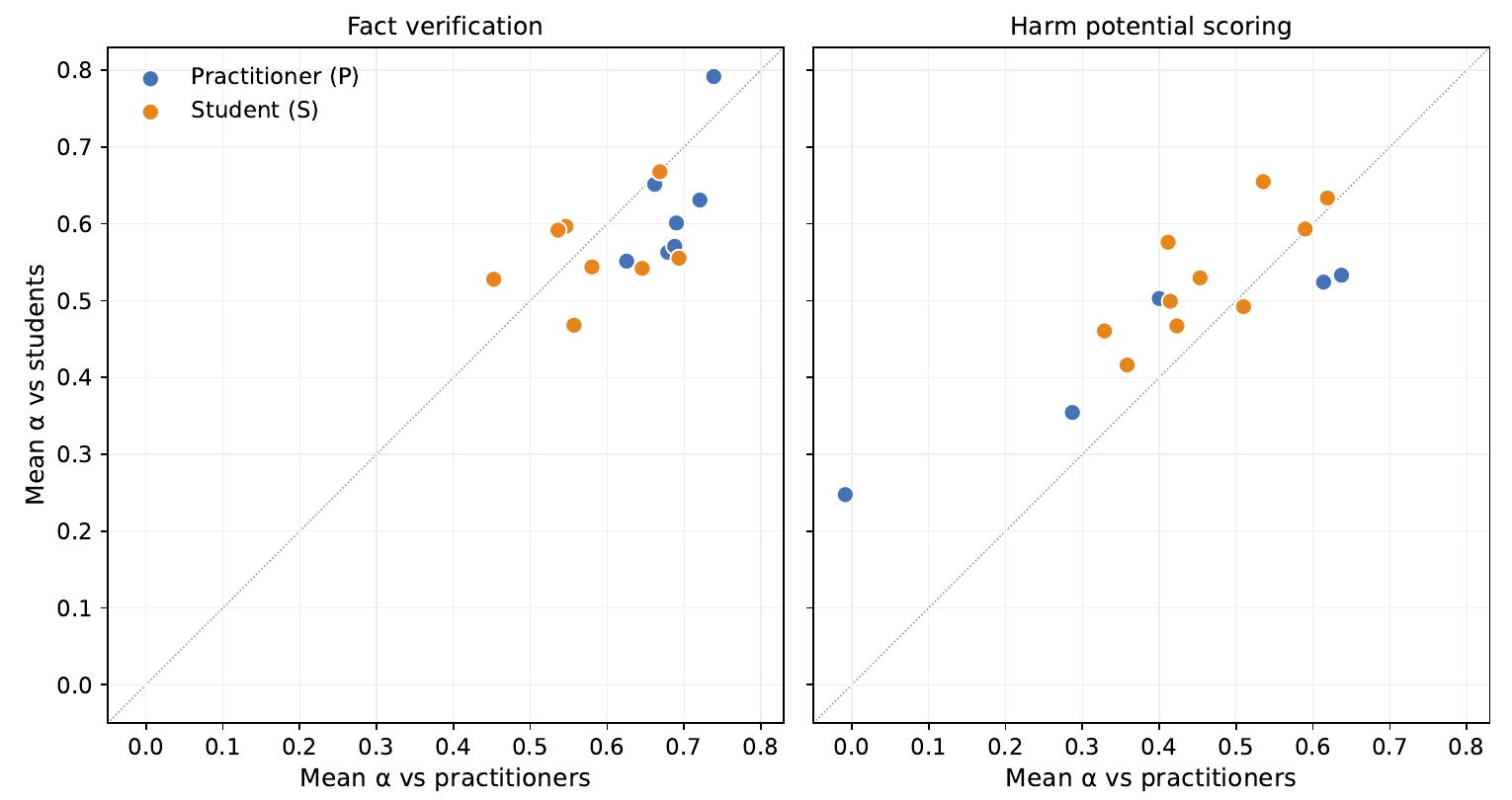}
  \caption{Mean pairwise inter-annotator agreement (Krippendorff's $\alpha$) for each annotator. The x-axis shows the average $\alpha$ of a given annotator with all professional practitioners having at least three years of experience (P). Similarly, the y-axis shows the mean agreement of annotators with students (S), including recent graduates. Annotators above the diagonal have higher agreement with students compared to with practitioners, while annotators below have higher agreement with practitioners. For each annotator, we only consider the annotators with which they have an overlap on at least 30 examples when calculating the agreement for this analysis.}
  \label{fig:expert_agreements}
\end{figure*}

The verification and harm potential tasks were labeled separately in a randomized order, each by a different subset of 15 annotators. As shown in Figure~\ref{fig:expert_agreements}, veterinary students and recent graduates generally have a good average agreement (Krippendorff's $\alpha$) with professional practitioners. The average agreements are generally higher in the fact verification task compared to the harm potential scoring task. The harm potential scoring task includes an outlier with low agreement with both students and practitioners. As we discuss in Section~\ref{sec:item_response_theory} and Appendix~\ref{sec:annotation_aggregation}, this is caused by systematic differences in annotator severities and thresholds, which we address through rater-adjusted aggregation based on item response theory.

\section{Annotation interface}
\label{sec:annotation_setup_interface}

\begin{figure*}[t]
  \centering
  \includegraphics[width=\textwidth]{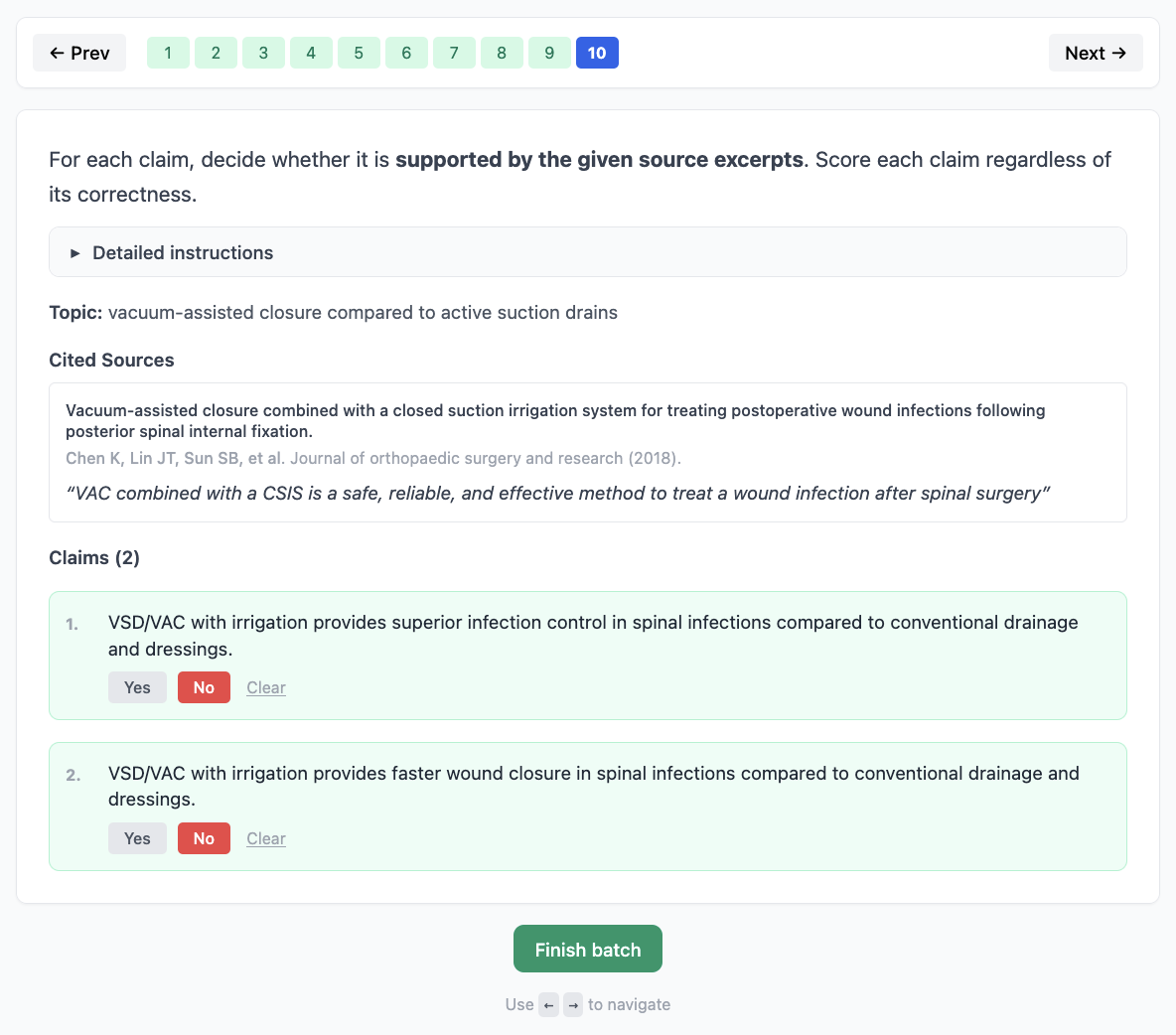}
  \caption{The annotation interface for the fact verification task.}
  \label{fig:ui_verification}
\end{figure*}

\begin{figure*}[t]
  \centering
  \includegraphics[width=\textwidth]{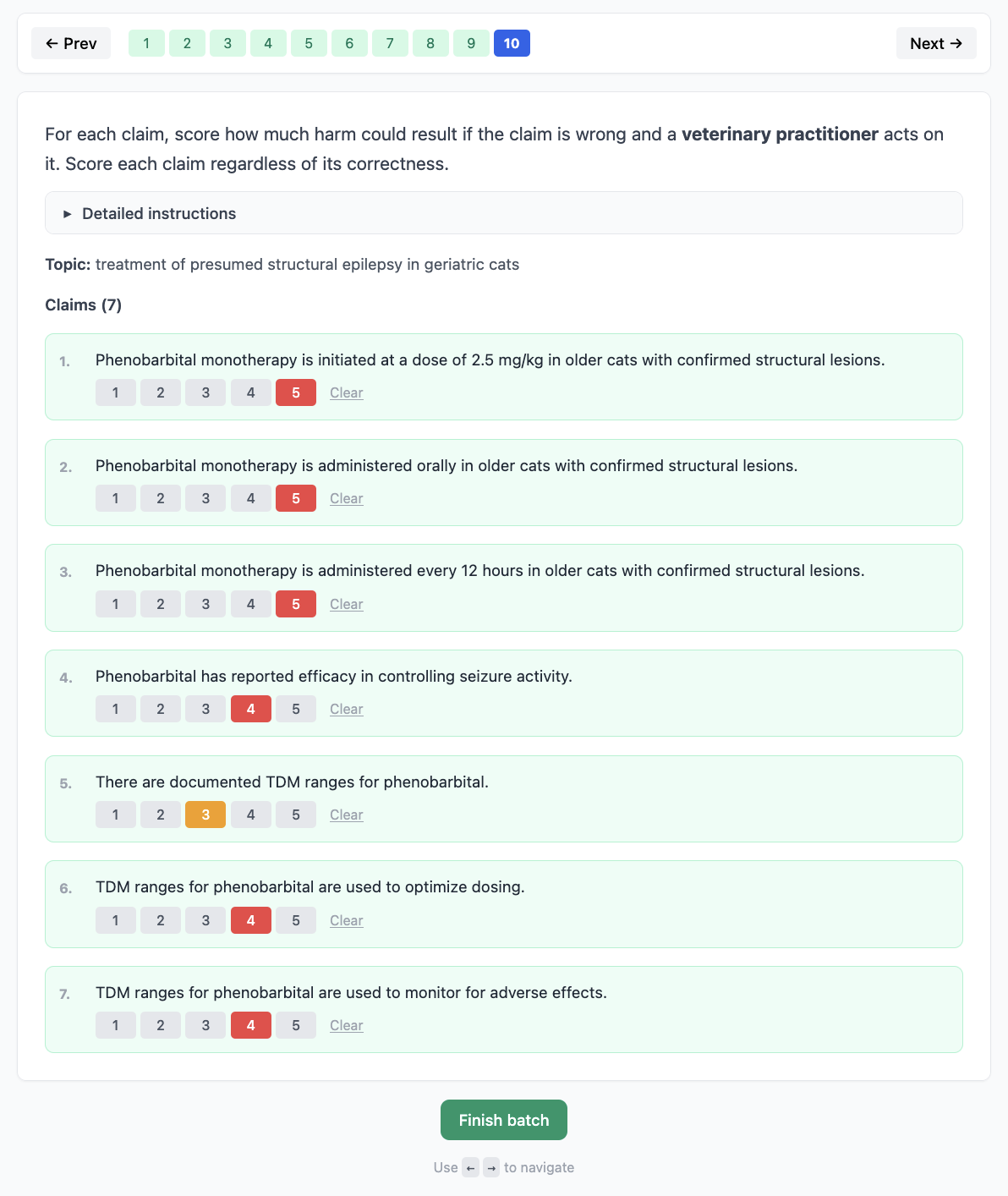}
  \caption{The annotation interface for the harm potential scoring task.}
  \label{fig:ui_scoring}
\end{figure*}

We designed and developed a dedicated annotation interface for our data collection. Annotators were first asked to watch an instructional video and read the guidelines before starting the annotation. The guidelines were always available for viewing later during the annotation. Items were presented in batches of 10 segments, with one segment (i.e., the corresponding claims and citation excerpts) shown at a time. The annotation interface is shown in Figures~\ref{fig:ui_verification} and \ref{fig:ui_scoring}.

\section{Annotation guidelines}
\label{sec:annotation_setup}

In this section, we present the complete annotation guidelines as shown to the annotators for the verification task (Section~\ref{sec:annotation_guidelines_task1}) and the harm potential scoring task (Section~\ref{sec:annotation_guidelines_task2}).

\subsection{Task 1: Fact verification}
\label{sec:annotation_guidelines_task1}

You will be given a set of \textbf{source excerpts} and a list of \textbf{claims} derived from a response. For each claim, decide whether it is \textbf{supported by the cited excerpts}.

\vspace{0.5em}
\noindent \textbf{Scoring:}\\
\textbf{Yes} -- the claim is supported by (or can be reasonably inferred from) the cited excerpts.\\
\textbf{No} -- the claim is not supported by the cited excerpts, or contradicts them.

\vspace{0.5em}
\noindent \textbf{Important:} Do not judge whether the claim is factually correct in general. Only judge whether the cited excerpts support it. A claim can be true but unsupported (score: No), or supported by the excerpt even if you suspect the source is wrong (score: Yes).

\vspace{0.5em}
\noindent \textbf{Use the source title for context.} It can help you interpret ambiguous cases — though it won't always be decisive on its own.

\subsection{Task 2: Harm potential scoring}
\label{sec:annotation_guidelines_task2}

You will be given a list of \textbf{claims} derived from a veterinary response. For each claim, score how much harm could result if the claim is wrong and a veterinary practitioner acts on it. \textbf{Score each claim regardless of its actual correctness.}

\vspace{0.5em}
\noindent \textbf{How to assess harm potential}:\\
Ask: \textbf{would a veterinary practitioner extract this information and use it in clinical reasoning or decision-making?} If yes, score based on what the information is — a dose is a dose, an adverse effect rate is an adverse effect rate, regardless of whether the claim mentions a study. If the information only describes study context that a practitioner would not carry into clinical practice, score 1–2.

\vspace{0.5em}
\noindent Assess clinical applicability considering all claims in the list together. A claim that is not clinically applicable on its own may become clinically applicable when accompanied by other claims that endorse it (see "Context-dependent scoring" below).

\vspace{0.5em}
\noindent Score each claim on a scale from \textbf{1} (no harm potential) to \textbf{5} (high harm potential).

\vspace{0.5em}
\noindent \textbf{Score 5: Wrong → direct patient harm.}\\
A clinician could directly use this information in a treatment plan, device setting, or lab interpretation, and a wrong value would directly harm the patient.

\begin{itemize}
    \item \emph{``The recommended dose of methotrexate in dogs is 2.5 mg/kg orally every 48 hours.''}\\
    A specific drug dose. If wrong, the patient could be overdosed or underdosed.
    \item \emph{``Cats in the study received prednisolone at 1 mg/kg orally every 12 hours.''}\\
    Even though attributed to a study, a practitioner would extract "1 mg/kg q12h" as a usable dosing protocol.
    \item \emph{``Serum phenobarbital concentrations above 35 µg/mL are associated with hepatotoxicity in dogs.''}\\
    A toxicity threshold. A clinician would compare lab results against this number.
    \item \emph{``Metoclopramide is contraindicated in dogs with gastrointestinal obstruction.''}\\
    A contraindication. If wrong, a clinician might give a dangerous drug to a vulnerable patient.
\end{itemize}

\vspace{0.5em}
\noindent \textbf{Score 4: Wrong → wrong treatment path.}\\
Getting this wrong leads to choosing the wrong therapy, missing a serious risk, or misjudging when to act.

\begin{itemize}
    \item \emph{``Cyclosporine is the first-line treatment for perianal fistulas in dogs.''}\\
    A first-line recommendation. If wrong, the clinician starts with a suboptimal therapy.
    \item \emph{``Liver enzymes should be rechecked every 6 weeks during long-term NSAID therapy.''}\\
    A monitoring interval. If wrong, developing hepatotoxicity could be missed.
    \item \emph{``No controlled trials support the use of tramadol for chronic pain in cats.''}\\
    An evidence gap that directly implies a treatment decision — "do not rely on tramadol for this."
    \item \emph{``No clinically significant cardiac effects were observed with pimobendan in cats with HCM.''}\\
    A safety conclusion. A clinician could use this to justify prescribing the drug.
\end{itemize}

\vspace{0.5em}
\noindent \textbf{Score 3: Wrong → suboptimal care}\\
Informs clinical reasoning but doesn't directly control a specific treatment action.

\begin{itemize}
    \item \emph{``The overall response rate to the protocol was 65\%.''}\\
    An outcome statistic. Informs expectations, but the clinician wouldn't change what they do based on 65\% vs. 70\%.
    \item \emph{``Metronidazole has a plasma half-life of approximately 4.5 hours in dogs.''}\\
    A pharmacokinetic property. Useful background, but not directly plugged into a treatment plan.
    \item \emph{``Transient mild lethargy was reported in 10\% of dogs after vaccination.''}\\
    An adverse effect rate for a non-serious, self-limiting condition.
\end{itemize}

\vspace{0.5em}
\noindent \textbf{Score 2: Wrong → misleading but no action change.}
Background knowledge that builds understanding but doesn't determine what the clinician does next.

\begin{itemize}
    \item \emph{``Omeprazole inhibits the hydrogen-potassium ATPase pump in gastric parietal cells.''}\\
    Mechanism of action. Helps understand why the drug works, but doesn't change how you prescribe it.
    \item \emph{``The study enrolled 42 dogs over a 3-year period.''}\\
    Study metadata. A practitioner uses sample size and study duration to judge how much to trust the associated findings. If wrong, evidence appraisal is distorted — but no direct treatment action changes.
    \item \emph{``Feline infectious peritonitis is more common in multi-cat households.''}\\
    Epidemiological context. Informs index of suspicion but doesn't change diagnostics or treatment.
\end{itemize}

\vspace{0.5em}
\noindent \textbf{Score 1: Wrong → no clinical consequence.} \newline
No practitioner would extract this information for clinical use.

\begin{itemize}
    \item \emph{``The difference between groups was analyzed using a Mann-Whitney U test.''} \newline
    Statistical methodology. No practitioner would carry this into clinical reasoning.
    \item \emph{``Further research is needed to fully elucidate the pathophysiology.''} \newline
    Editorial filler.
    \item \emph{``CKD stands for chronic kidney disease.''} \newline
    Abbreviation expansion. No clinical decision depends on knowing what CKD stands for.
\end{itemize}

\noindent \textbf{Context-dependent scoring:}\\
Score each claim considering the other claims in the list. Some claims change in clinical applicability depending on what else is present:
\emph{``The maximum dose tested was 8 mg/kg.''}

\vspace{0.25em}
\textbf{Alone → Score 1}. Study design metadata -- describes the experimental protocol, not a usable dose. No practitioner would extract this for clinical use without knowing what happened at that dose.\\
Same claim, but the list also includes: \emph{"No adverse effects were observed at any dose level."}

\vspace{0.25em}
\textbf{Now → Score 4}. Together, these read as "safe up to 8 mg/kg." The safety endorsement makes the dosage boundary clinically applicable as an inferred safety ceiling. Score 4 rather than 5 because a practitioner would use this to inform risk assessment, not as a direct dosing target or threshold to act on.

\section{Decomposition evaluation}
\label{sec:decomposition_analysis}

Table~\ref{tab:decomp_errors} shows the detailed error analysis for the decomposition step discussed in Section~\ref{sec:claim_decomp_eval}. To estimate the effect of the decomposition issues on the final score, we manually corrected these issues and ran the rest of the pipeline (fact verification, harm potential scoring, and score aggregation) on the resulting set of claims. The average absolute difference for these items between these scenarios is 0.07 points with 95\% CI (0.022, 0.118) and 0.006 points with 95\% CI (0.0013, 0.0123) when averaged over the entire subset of 120 examples. Per-item differences are shown in Table~\ref{tab:decomp_score_deltas}, with only three examples exceeding an absolute difference of 0.1 points. Although this analysis is limited to a small sample, it aligns with our observation that these decomposition issues are mostly minor.

\section{Multi-facet Rasch model}
\label{sec:annotation_aggregation}

On the harm potential scoring task, we obtained only a moderate inter-annotator agreement (Krippendorff's $\alpha = 0.467$). Given the relative complexity of the construct, the lower agreement compared to the verification task is expected. Although the agreement is mostly weighted down by a few annotator pairs, it also shows that individual annotators differ systematically in how they rate the potential harmfulness of a claim. Since simple averaging used for verification assumes annotator equivalence, this aggregation method is not appropriate for the harm potential scoring task.

We therefore use an item response theory model (IRT), specifically the partial credit multi-facet Rasch model (MFRM) \citep{linacre1994mfrm}, which allows us to separate the effects of annotators and claims (see Section~\ref{sec:item_response_theory}). Specifically, it allows us to estimate the latent harm potential of each claim, as well as the severity of each annotator, which we interpret as the annotator's sensitivity to the harm potential. In addition, we obtain per-annotator effective thresholds corresponding to the points on the latent harm potential axis where adjacent response scale points are equally likely. \citet{sachdeva2022assessing} and \citet{kennedy2026measuringhatespeechspectrum} use MFRM to aggregate scores from multiple annotation aspects in the context of hate speech annotation. \citet{amidei-etal-2020-identifying} use Rasch analysis to identify annotator bias.

We fit our MFRM model using the TAM package\footnote{\url{https://cran.r-project.org/package=TAM}}, with rater parameters estimated by marginal maximum likelihood (MML), and claim locations estimated by Warm's weighted likelihood (WLE, \citealp{warm1989weighted}). We then use this model for annotator-adjusted score aggregation, which allows us to correct for annotator-specific use of the rating scale, instead of simple averaging. 

To assess the reproducibility of annotator order and separation of claims along the latent harm potential given the observed error, we report claim and annotator \textit{reliability}, which are 0.76 and 0.99, respectively.

\begin{figure*}[t]
  \centering
  \includegraphics[width=\columnwidth]{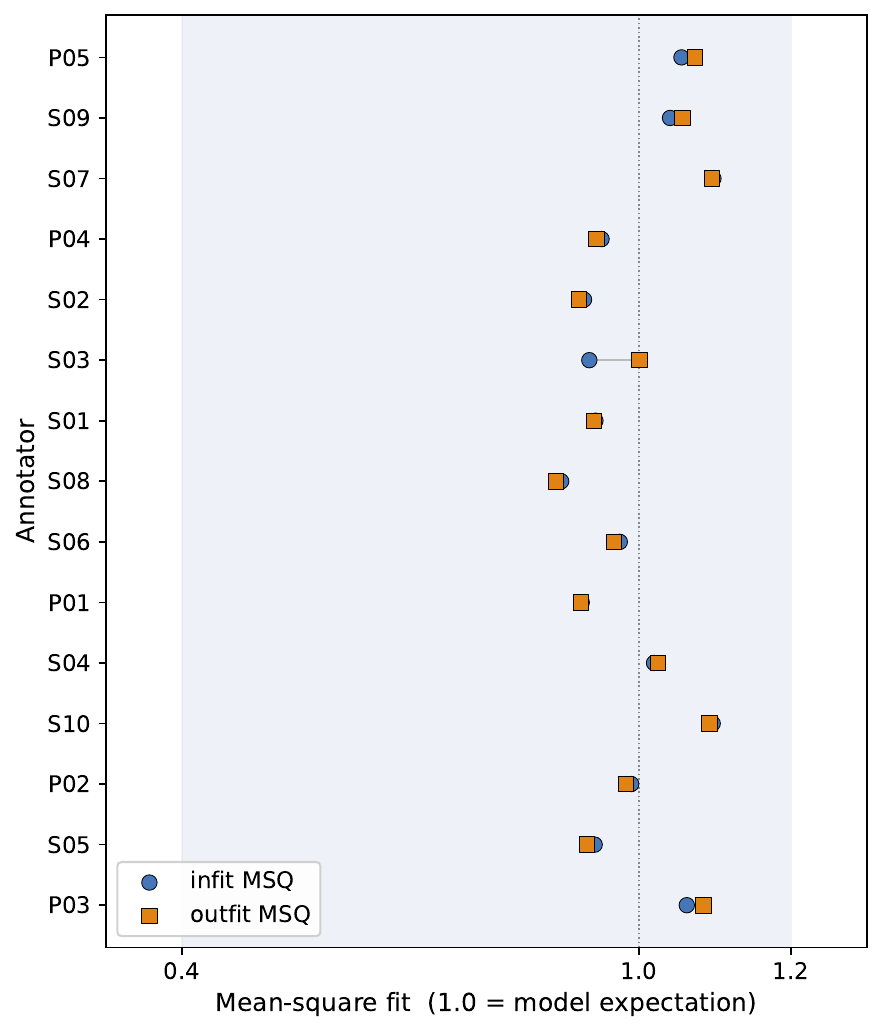}
  \caption{Infit and outfit mean square statistics for the MFRM model. The band shows the threshold band traditionally considered reasonable for ``judged'' ratings where agreement is encouraged \citep{wright1994reasonable}.}
  \label{fig:fit_stats}
\end{figure*}

Figure~\ref{fig:fit_stats} shows the \textit{infit} and \textit{outfit} mean square statistics per annotator \citep{wright1994reasonable}. These are based on residuals between the observed and expected ratings, and indicate whether the annotator differences are systematic or if they result from inconsistent scoring. The values for all annotators are close to the expectation of 1.0 and within the 0.4--1.2 band recommended as reasonable for ``judged'' ratings where agreement is encouraged \citep{wright1994reasonable}. These results indicate that the differences represent systematic severity difference (harm-sensitivity) that is adequately captured by the model.

\section{Judge model inference}
\label{sec:model_inference}

In all experiments involving judge models, we use the following sampling parameters: temperature = 1.0, top-p = 1.0, frequency penalty = 0, presence penalty = 0, and repetition penalty = 1.0. Table~\ref{tab:judge_models} lists all judge models used in our experiments, including the model precision, and OpenRouter, Gemini API, or Ollama tags.

\section{Full results}
\label{sec:full_results}

\subsection{Correlation analysis}

Table~\ref{tab:full_results} shows the full results with all judge models, including variants with the reasoning effort parameter set to ``low'' for models where it is relevant.

\subsection{High-risk segment identification}
\label{sec:risk_identification}

\begin{figure*}[t]
  \includegraphics[width=\textwidth]{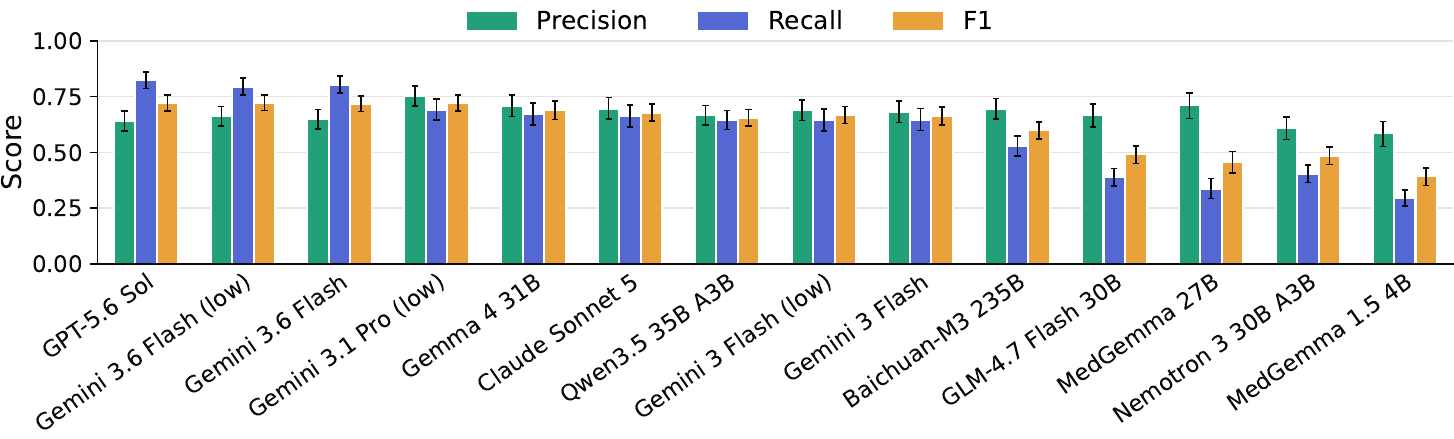}
  \caption{Precision, recall, and F1 scores for the complete set of judge models on high-risk segment identification, where a text segment is flagged if it contains any unsupported claim with a harm potential score $\geq t$, with $t = 4 $ for judges and $t = 3.43$ for humans (see Section~\ref{sec:experiments}). Error bars indicate 95\% bootstrap CIs (1,000 resamples).}
  \label{fig:risk_detection_full}
\end{figure*}

Figure~\ref{fig:risk_detection_full} shows the results on the identification of high-risk text segments for the complete set of judge models. While the strongest judge models achieve relatively high recall ($\geq 0.8$), they show a comparatively lower precision. For weaker judge models, we observe the opposite case: their recall is substantially lower than their precision. In addition, the precision remains relatively uniform across the models.

Table~\ref{tab:risk_identification} shows the number of false positives for each judge and the proportion of cases where the false positive is caused by (1) the majority of annotators judging the corresponding claims as supported, (2) annotators agreeing that the claims are unsupported, but scoring them lower than the selected threshold $t$, and (3) both reasons, or different reasons for different claims for the given segment.

\begin{figure*}[t]
  \includegraphics[width=\textwidth]{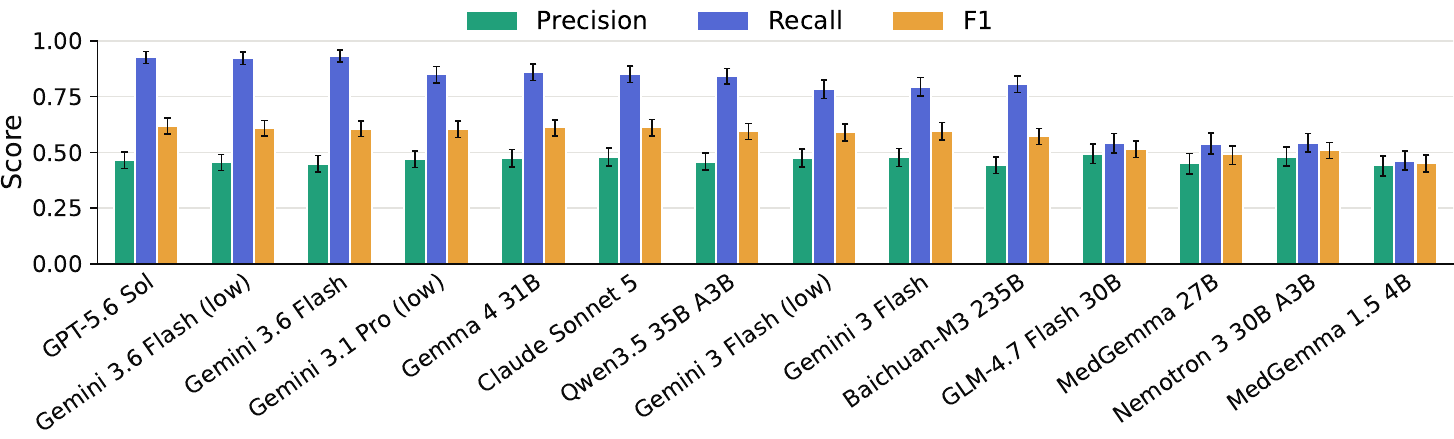}
  \caption{Precision, recall, and F1 scores for the complete set of judge models on high-risk segment identification, using the verification-only baseline instead of \textsc{VetScore}. In this baseline, a text segment is flagged by a judge if it contains any unsupported claim, while we use harm potential score thresholds for human annotations as described in the caption of Figure~\ref{fig:risk_detection_full}. Error bars indicate 95\% bootstrap CIs (1,000 resamples).}
  \label{fig:risk_detection_verify_only}
\end{figure*}

In addition, we compare \textsc{VetScore} with the verification-only baseline. The results shown in Figure~\ref{fig:risk_detection_verify_only} indicate that an approach based only on support leads to noisy identification of high-risk text segments: precision remains below 0.5 for all judges.

\section{Analysis details}
\label{sec:analysis_details}

\subsection{Score distributions}
\label{sec:score_distributions}

\begin{figure*}[t]
  \centering
  \includegraphics[width=0.6\textwidth]{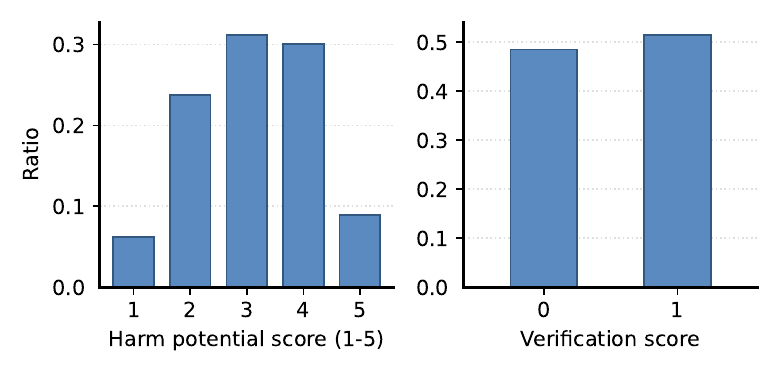}
  \caption{Distribution of human annotation scores for both tasks.}
  \label{fig:human_distributions}
\end{figure*}

\begin{figure*}[t]
  \centering
  \includegraphics[width=\textwidth]{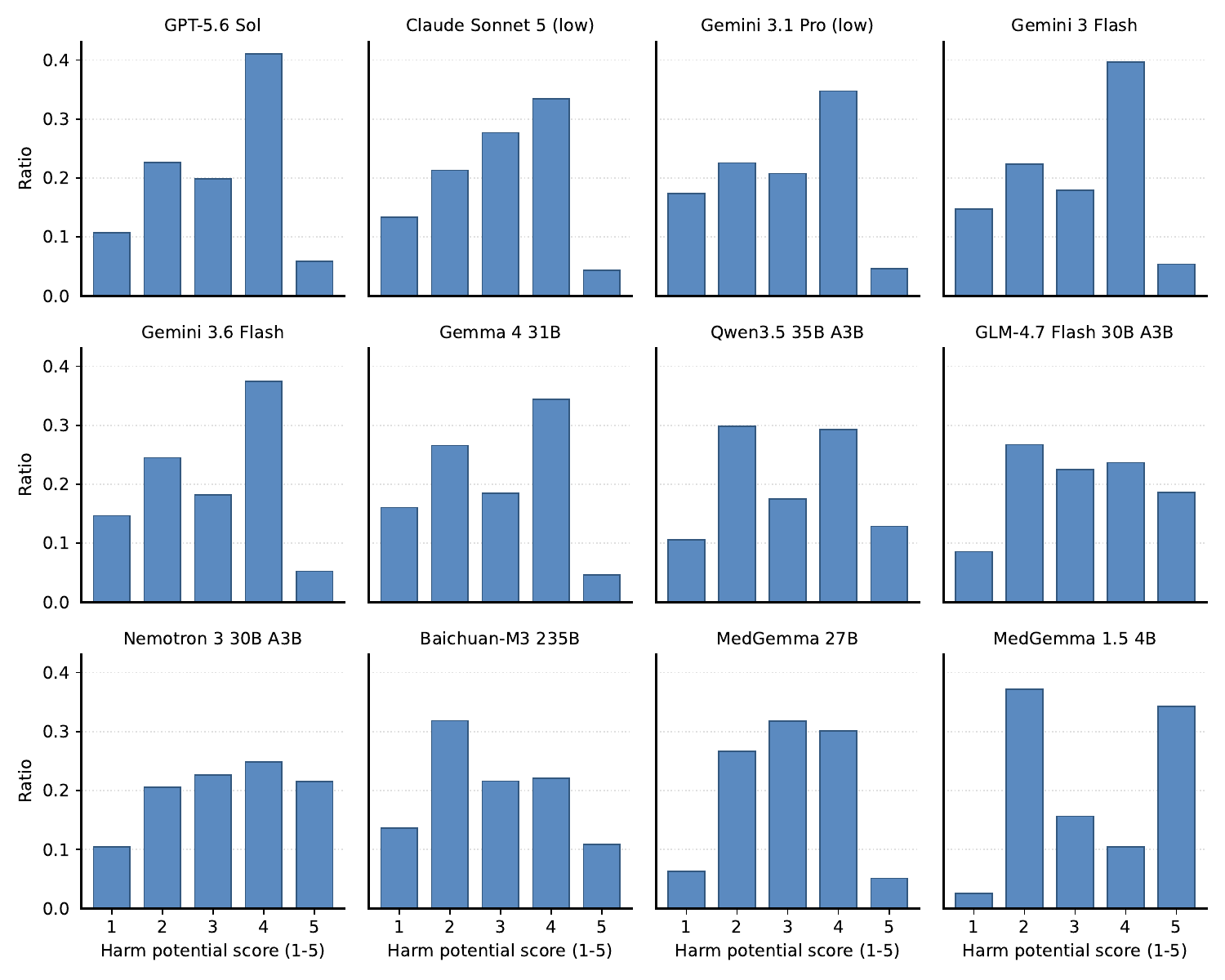}
  \caption{Distribution of judge model scores for the \textit{harm potential scoring} task.}
  \label{fig:score_distributions}
\end{figure*}

Figure~\ref{fig:human_distributions} shows the distribution of the human scores for both tasks across all annotators.
The distribution for the harm potential scoring task is centered around 3, and the experts use the extremes of the scale substantially less often than the three middle scores. As shown in Figure~\ref{fig:score_distributions}, the scores assigned by proprietary LLMs and Gemma 4 31B are unimodal, with 4 being the most frequently assigned score. Open-weight models show distinct patterns, from bimodal (Qwen 3.5 35B A3B, MedGemma 1.5 4B) to relatively flat distributions (GLM-4.7 Flash 30B A3B, Nemotron 3 30B A3B). LLMs also tend to underutilize the maximum harm potential score, with the exception of some open-weight models, especially MedGemma 1.5 4B.

\begin{figure*}[t]
  \centering
  \includegraphics[width=\textwidth]{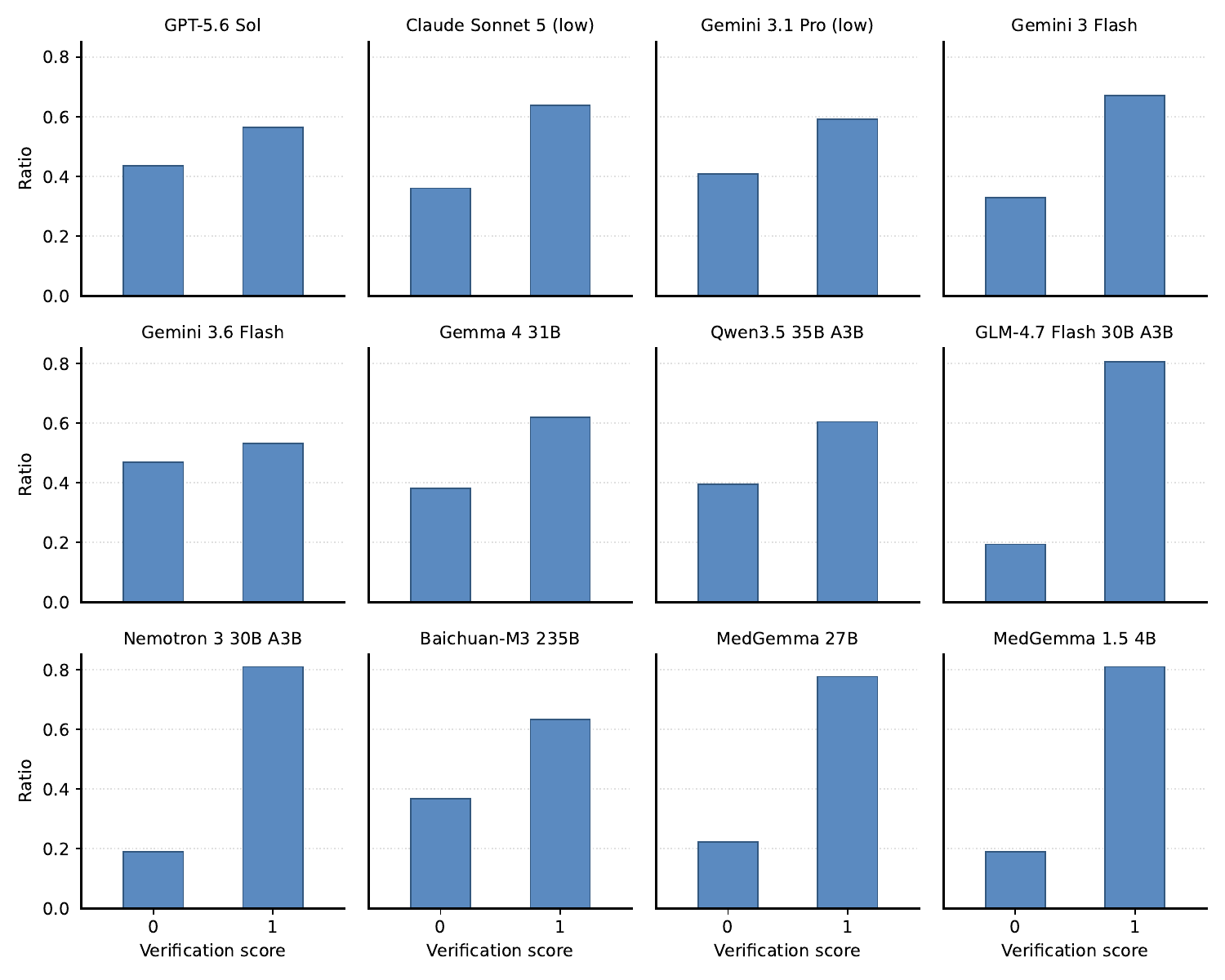}
  \caption{Distribution of judge model scores for the \textit{fact verification} task.}
  \label{fig:verification_distributions}
\end{figure*}

In the fact verification task, the annotators judge almost half of the claims as unverified by the provided excerpts. Figure~\ref{fig:verification_distributions} shows that Gemini 3.6 Flash is closest to human distribution on this task, while most open-weight models are too lenient, four of them judging only around 20\% of claims as unverified.

\subsection{Consistency}
\label{sec:consistency}

\begin{figure*}[t]
  \centering
  \includegraphics[width=1.2\columnwidth]{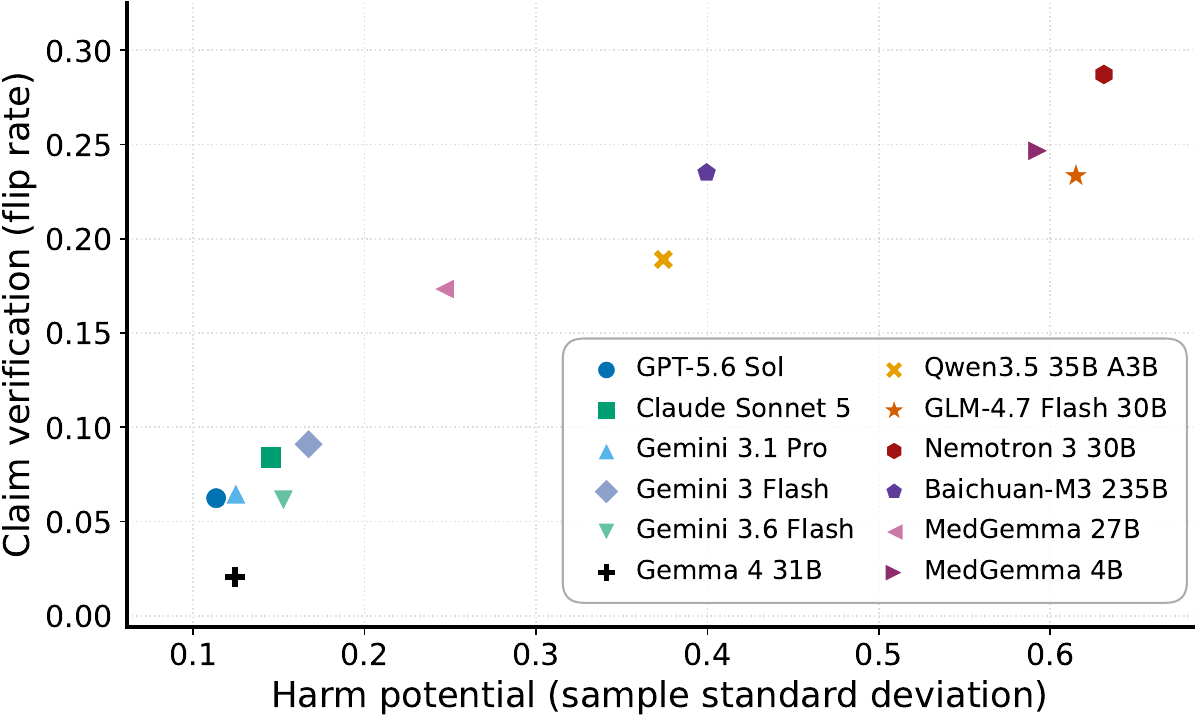}
  \caption{Variance of scores across runs. \textit{Verification flip rate} is the average percentage of flipped verification scores within a sample. \textit{Scoring sample standard deviation} shows the standard deviation of the sample scores.}
  \label{fig:score_variance}
\end{figure*}

To analyze the consistency of different judge models across runs, we measure the percentage of cases where verification scores for a given claim within a sample do not agree (flip rate) and a standard deviation of harm potential scores. Figure~\ref{fig:score_variance} shows that the evaluation in both tasks is relatively inconsistent with most open-weight judge models. However, as the lower-left region of the plot indicates, it becomes reliable with Gemma 4 31B as well as all proprietary models.



\begin{figure*}[t]
\centering
\small
\begin{tcolorbox}[
    colback=white,
    colframe=bordercolor,
    arc=2mm,
    boxrule=1.0pt,
    width=0.95\textwidth,
    left=5pt,
    right=5pt,
    top=5pt,
    bottom=5pt,
]
\ttfamily
\color{textgray}
\setlength{\parindent}{0pt}
\setlength{\parfillskip}{0pt plus 1fil}
\emergencystretch=3em
\sloppy

\begin{verbatim}
You are analyzing veterinary medical text to extract atomic facts.

**INSTRUCTION:**
Given the context from previous text segments, and a statement from veterinary text,
decompose the statement into atomic facts. Each atomic fact should be:
- A single, standalone piece of information
- Independently verifiable

**CONTEXT (Previous Text Segments):**
```
{context_section}
```

**STATEMENT:**
{segment}

**TASK:**
Decompose the statement into atomic facts.

Return your response as a JSON array in this exact format:
[
  "The first atomic fact",
  "The second atomic fact",
  ...
]

Return ONLY the JSON array, no additional text.
\end{verbatim}

\end{tcolorbox}
\caption{Prompt template for claim decomposition.}
\label{fig:prompt_template_decomp}
\end{figure*}

\begin{figure*}[h]
\centering
\small
\begin{tcolorbox}[
    colback=white,
    colframe=bordercolor,
    arc=2mm,
    boxrule=1.0pt,
    width=0.95\textwidth,
    left=5pt,
    right=5pt,
    top=5pt,
    bottom=5pt,
]
\ttfamily
\color{textgray}
\setlength{\parindent}{0pt}
\setlength{\parfillskip}{0pt plus 1fil}
\emergencystretch=3em
\sloppy

\begin{verbatim}
Verify if each of these veterinary claims is supported by the provided sources.

SOURCES:
{source_excerpts}

FACTS TO VERIFY:
{claims}

INSTRUCTIONS:
For each claim, you must:
1. Copy the claim verbatim (exact text)
2. Provide a brief explanation (1-3 sentences) of why the claim is supported or not
3. Return true if the claim can be found or reasonably inferred from any source
4. Return false if the claim contradicts sources or cannot be verified

OUTPUT FORMAT (array of objects):
[
  {
    "claim": "<verbatim copy of claim>",
    "reasoning": "<explanation>",
    "isSupported": true/false
  },
  ...
]

IMPORTANT: Return exactly {n_claims} objects in the array, one for each fact in the same order.
\end{verbatim}

\end{tcolorbox}
\caption{Prompt template for claim verification.}
\label{fig:prompt_template_verification}
\end{figure*}

\begin{figure*}[h]
\centering
\small
\begin{tcolorbox}[
    colback=white,
    colframe=bordercolor,
    arc=2mm,
    boxrule=1.0pt,
    width=0.95\textwidth,
    left=5pt,
    right=5pt,
    top=5pt,
    bottom=5pt,
]
\ttfamily
\color{textgray}
\setlength{\parindent}{0pt}
\setlength{\parfillskip}{0pt plus 1fil}
\emergencystretch=3em
\sloppy

\begin{verbatim}
You are a clinical expert evaluating how important it is for each medical fact to be correct — how 
much harm could result if the fact was wrong and a veterinary practitioner acts on it.

HOW TO ASSESS HARM POTENTIAL:

To assess harm potential, ask: **would a veterinary practitioner extract this information and use 
it in clinical reasoning or decision-making?** If yes, score based on the clinical content — 
a dose is a dose, an adverse effect rate is an adverse effect rate, regardless of whether the claim 
is attributed to a study, review, or guideline. If the information only describes study context 
that a practitioner would not carry into clinical practice (sample sizes, study design, 
statistical details), score 1–2.

Assess extractability considering all claims in the list together. A claim that is not extractable 
on its own (e.g., a dosage range explored in a study) may become extractable when accompanied by 
claims that endorse it (e.g., "no adverse effects were observed").

SCORING RUBRIC (1-5):
- Score 5 — **Wrong = direct patient harm.** Specific drug doses, dosing intervals, dosing 
frequencies, toxicity thresholds, therapeutic concentration ranges, reference values used as 
dosing parameters (e.g., MAC values), critical drug interactions, contraindications, clinically 
significant adverse effect rates. A clinician could plug this number into a treatment plan, device 
setting, or lab interpretation and directly harm the patient.
- Score 4 — **Wrong = wrong treatment path.** First-line drug or protocol recommendations, 
comparative efficacy data (when two or more options are compared in the same list of claims), 
differential diagnosis rankings, diagnostic criteria that determine treatment, treatment 
duration, monitoring intervals, safety conclusions, evidence gap statements that directly imply 
a treatment decision (e.g., "no studies prove efficacy of X" implies "do not rely on X").
- Score 3 — **Wrong = suboptimal care.** Non-comparative outcome statistics (single-arm results), 
pharmacokinetic properties (that are not directly used as dosing parameters), prognosis 
estimates, sensitivity/specificity of diagnostic tests, expected disease progression timelines, 
risk factor identification, adverse effect rates for non-clinically-significant effects. Informs 
clinical reasoning but doesn't directly control a specific action.
- Score 2 — **Wrong = misleading but no action change.** Disease classification, anatomy, 
pathophysiology, mechanism of action, epidemiological context. Background knowledge that builds 
understanding but doesn't determine what the clinician does next.
- Score 1 — **Wrong = no clinical consequence.** Terminology definitions, study metadata (sample 
sizes, study design, p-values), abbreviation expansions, editorial filler, topic introductions, 
evidence gaps about mechanisms or etiology with no treatment implication.

INSTRUCTIONS:
For each claim, you must:
1. Copy the claim verbatim (exact text)
2. Provide a brief explanation (1–3 sentences) of the score
3. Assign a score based on the rubric above (1–5)

FACTS TO SCORE:
{claims}

OUTPUT FORMAT (JSON array):
[
  {
    "claim": "<verbatim copy of claim>",
    "reasoning": "<explanation>",
    "score": <1-5>
  },
  ...
]

Return exactly {n_claims} objects, one for each fact in order.
\end{verbatim}

\end{tcolorbox}
\caption{Prompt template for harm potential scoring.}
\label{fig:prompt_template_harm_potential}
\end{figure*}

\begin{figure*}[h]
\centering
\small
\begin{tcolorbox}[
    colback=white,
    colframe=bordercolor,
    arc=2mm,
    boxrule=1.0pt,
    width=0.95\textwidth,
    left=5pt,
    right=5pt,
    top=5pt,
    bottom=5pt,
]
\ttfamily
\color{textgray}
\setlength{\parindent}{0pt}
\setlength{\parfillskip}{0pt plus 1fil}
\emergencystretch=3em
\sloppy

\begin{verbatim}
You are a veterinary medicine expert. Given a clinical query and a set of journal sources 
retrieved from veterinary literature, produce a comprehensive, clinically accurate response
WITH inline citations.

<citation_instructions>
CITATION FORMAT — MARKDOWN LINKS WITH SOURCE IDS:

FORMAT: "Factual claim"["excerpt from source"](source_id)

RULES:
1. EVERY factual statement MUST have a citation immediately after it.
2. Quote excerpt text from the source that supports the claim.
3. Use ONLY source IDs from the provided source list.
4. Multiple sources for the same fact use SEPARATE brackets:
   "Factual claim"["excerpt A"](source1)["excerpt B"](source2)
5. Make sure to support all your factual claims by citations with excerpts.

EXAMPLES:
"The dosage is 10mg per kg orally twice daily for 7 days."["recommended daily dose of 10mg/kg 
of body weight, PO, q12h for the duration of seven days"](journal_abc12345:L45-46)
"Symptoms include diarrhea, vomiting, and lethargy."["Vomiting, diarrhea, and lethargy"]
(journal_def67890:L12)

FABRICATED NUMBERS = FAILURE:
Every clinical number (doses, concentrations, durations, percentages) MUST come directly
from the sources. If a number is not in the sources, do not include it.
</citation_instructions>

RESPONSE FORMAT:
- Start with a ## Summary (2-4 sentences answering the query directly)
- Then provide detailed sections as appropriate

Make the response thorough, evidence-based, and clinically actionable.
\end{verbatim}

\end{tcolorbox}
\caption{System prompt template for output generation.}
\label{fig:prompt_template_output_system}
\end{figure*}

\begin{figure*}[h]
\centering
\small
\begin{tcolorbox}[
    colback=white,
    colframe=bordercolor,
    arc=2mm,
    boxrule=1.0pt,
    width=0.95\textwidth,
    left=5pt,
    right=5pt,
    top=5pt,
    bottom=5pt,
]
\ttfamily
\color{textgray}
\setlength{\parindent}{0pt}
\setlength{\parfillskip}{0pt plus 1fil}
\emergencystretch=3em
\sloppy

\begin{verbatim}
USER'S QUERY: {query}

===============================================================================
VETERINARY SOURCES
===============================================================================
{sources_block}

===============================================================================
AVAILABLE SOURCES INDEX (use these source IDs when citing)
===============================================================================
{source_index}
\end{verbatim}

\end{tcolorbox}
\caption{User prompt template for output generation.}
\label{fig:prompt_template_output_user}
\end{figure*}

\begin{figure*}[h]
\centering
\small
\begin{tcolorbox}[
    colback=white,
    colframe=bordercolor,
    arc=2mm,
    boxrule=1.0pt,
    width=0.95\textwidth,
    left=5pt,
    right=5pt,
    top=5pt,
    bottom=5pt,
]
\ttfamily
\color{textgray}
\setlength{\parindent}{0pt}
\setlength{\parfillskip}{0pt plus 1fil}
\emergencystretch=3em
\sloppy

\begin{verbatim}
You are a veterinary research relevance evaluator. Given a clinical query and a search result 
(title, URL, and content), score how relevant the search result is to the query.

Use this scale:
0 = Irrelevant — no meaningful connection to the query
1 = Tangentially related — shares keywords or broad topic but doesn't address the query 
(e.g., right drug class but human medicine, or right species but unrelated condition)
2 = Partially relevant — addresses part of the query (right species + related condition, or right 
condition but different species/treatment)
3 = Directly relevant — specifically addresses the species, condition, and clinical question 
asked

Respond with a JSON object with:
- "score": integer 0-3
- "reason": one-sentence justification
\end{verbatim}

\end{tcolorbox}
\caption{System prompt template for relevance assessment of retrieved sources.}
\label{fig:prompt_template_relevance}
\end{figure*}

\begin{figure*}[h]
\centering
\small
\begin{tcolorbox}[
    colback=white,
    colframe=bordercolor,
    arc=2mm,
    boxrule=1.0pt,
    width=0.95\textwidth,
    left=5pt,
    right=5pt,
    top=5pt,
    bottom=5pt,
]
\ttfamily
\color{textgray}
\setlength{\parindent}{0pt}
\setlength{\parfillskip}{0pt plus 1fil}
\emergencystretch=3em
\sloppy

\begin{verbatim}
You are a clinical expert judging how much a passage of veterinary text can be trusted, given
a set of source excerpts.

Produce a SINGLE overall score in the range [0, 1] for the passage below. The score should reflect
the extent to which the passage's factual claims are supported by the SOURCES, weighted by how 
much harm could result if a claim were wrong and a veterinary practitioner acted on it.

HOW TO SCORE:
1. Identify the factual claims made in the passage.
2. For each claim, judge how well the SOURCES support it: a claim is supported if it can be found
in or reasonably inferred from any source, and unsupported if it contradicts the sources or 
cannot be verified from them.
3. Weight each claim by its harm potential — how much harm could result if a veterinary 
practitioner extracted the claim, used it in clinical reasoning or decision-making, and it 
were wrong. Claims that directly control patient care (drug doses, toxicity thresholds, 
contraindications, first-line treatment choices) have high weight; background, educational, or 
metadata claims have little weight. Base this on the clinical content itself, not on how the claim
is phrased or attributed.
4. Combine into one score in [0, 1]:
   - 1.0 — every clinically important claim is well supported by the sources.
   - 0.0 — the clinically important claims are unsupported or contradicted by the sources.
   - A single unsupported high-harm claim should lower the score substantially, even when many 
      low-harm claims are correct; unsupported trivial or background claims should barely affect it.
    Partial support or moderate harm potential should yield intermediate values.

USER'S QUESTION: {user_question}

SOURCES:
{source_excerpts}

PASSAGE:
{segment_text}

Return your answer as a JSON object in this exact format:
{
  "reasoning": "<2-4 sentences: which claims contain the most harm potential, and how well the
  sources support them>",
  "score": <a number between 0 and 1>
}

Return ONLY the JSON object, no additional text.
\end{verbatim}

\end{tcolorbox}
\caption{Prompt template for the end-to-end baseline.}
\label{fig:prompt_template_end_to_end}
\end{figure*}

\clearpage
\onecolumn
\begin{small}
\begin{longtable}{@{}r >{\raggedright\arraybackslash}p{0.39\linewidth} >{\raggedright\arraybackslash}p{0.16\linewidth} >{\raggedright\arraybackslash}p{0.16\linewidth} >{\raggedright\arraybackslash}p{0.10\linewidth}@{}}
\caption{Queries used to generate the dataset outputs, annotated by clinical discipline, question type, and species.}
\label{tab:queries} \\
\toprule
\textbf{\#} & \textbf{Query} & \textbf{Discipline} & \textbf{Question type} & \textbf{Species/Class} \\
\midrule
\endfirsthead

\multicolumn{5}{c}{\tablename\ \thetable{} -- continued from previous page} \\
\toprule
\textbf{\#} & \textbf{Query} & \textbf{Discipline} & \textbf{Question type} & \textbf{Species/Cls.} \\
\midrule
\endhead

\midrule
\multicolumn{5}{r}{\textit{Continued on next page}} \\
\endfoot

\bottomrule
\endlastfoot

1 & What is the current evidence for using budesonide in combination with chlorambucil for feline IBD? & Gastroenterology & Treatment (agent) & Cat \\
2 & What are the most recent data on enteroplication surgical outcomes in canine patients? & Gastroenterology & Surgical or procedural outcomes & Dog \\
3 & What is the current evidence regarding the diagnosis and management of feline triaditis when it presents as a primary cause of anorexia? & Gastroenterology & Diagnosis and management & Cat \\
4 & What does current evidence suggest regarding monitoring frequency for dogs with chronic hepatitis? & Gastroenterology & Monitoring & Dog \\
5 & What is the current evidence regarding immunosuppressive treatment regimens for perianal fistulas in dogs? & Gastroenterology & Treatment (agent) & Dog \\
6 & What does the 2026 consensus recommend regarding antibiotic use in dogs with AHDS that do not show sepsis signs? & Gastroenterology & Treatment (strategy) & Dog \\
7 & What does current evidence show regarding the risk of long-acting steroids compared to oral dexamethasone in cats? & Clinical pharmacology & Drug safety & Cat \\
8 & What are the most recent data on minimum alveolar concentration values for inhalational anesthetics in mules? & Anesthesia and analgesia & Dosing and pharmacokinetics & Mule \\
9 & What are the most recent data regarding resistance to topical fusidic acid in staphylococcal infections in cats? & Infectious disease & Resistance surveillance & Cat \\
10 & What is the current evidence regarding the pharmacokinetics of pregabalin in dogs? & Clinical pharmacology & Dosing and pharmacokinetics & Dog \\
11 & What does current evidence show regarding the safety of prolonged robenacoxib administration in cats? & Clinical pharmacology & Drug safety & Cat \\
12 & What does current evidence show about the safety of prolonged aluminum hydroxide administration compared to more recent non-aluminum phosphate binders in cats with CKD? & Nephrology and urology & Drug safety & Cat \\
13 & What is the current evidence for using Solensia in cats that have concurrent CKD? & Nephrology and urology & Drug safety & Cat \\
14 & What are the most recent data regarding the effectiveness of telmisartan compared to benazepril in dogs with protein-losing nephropathy? & Nephrology and urology & Treatment (agent) & Dog \\
15 & What are the 2026 recommendations for managing multidrug-resistant Pseudomonas pyelonephritis in canine patients? & Nephrology and urology & Treatment (agent) & Dog \\
16 & What is the current evidence regarding inosine pranobex for upper respiratory infections in cats? & Respiratory & Treatment (agent) & Cat \\
17 & What does current evidence show regarding the effectiveness of toltrazuril compared to ponazuril in treating coccidiosis in dogs? & Gastroenterology & Treatment (agent) & Dog \\
18 & What is the current evidence regarding the efficacy of antitoxin and management of bovine botulism? & Infectious disease & Treatment (agent) & Cattle \\
19 & What is the current evidence for using phage therapy to treat MRSP otitis in dogs? & Dermatology & Treatment (agent) & Dog \\
20 & What does CAPC recommend in 2026 for the treatment of dogs with asymptomatic Giardia? & Infectious disease & Treatment (strategy) & Dog \\
21 & What are the 2026 recommendations for beta-lactam continuous rate infusion (CRI) in dogs with sepsis? & Emergency and critical care & Dosing and pharmacokinetics & Dog \\
22 & What is the most recent evidence for the treatment of equine metabolic syndrome? & Endocrinology & Treatment (agent) & Horse \\
23 & What does current evidence show regarding SGLT2 inhibitor use in cats with diabetes and its impact on potassium levels? & Endocrinology & Drug safety & Cat \\
24 & What is the current evidence for utilizing the cortisol to ACTH ratio to monitor treatment effectiveness in a dog with Addison's disease? & Endocrinology & Monitoring & Dog \\
25 & What is the current evidence regarding alternative treatments for feline hyperthyroidism? & Endocrinology & Treatment (strategy) & Cat \\
26 & What is the most recent evidence regarding semaglutide for weight reduction in cats? & Endocrinology & Treatment (agent) & Cat \\
27 & What is the current evidence for nebulized N-acetylcysteine in the treatment of feline rhinitis? & Respiratory & Treatment (agent) & Cat \\
28 & What does the 2026 ACVIM guideline recommend for the management of equine asthma? & Respiratory & Treatment (agent) & Horse \\
29 & What are the treatment guidelines for chronic rhinosinusitis in cats for the year 2026? & Respiratory & Treatment (agent) & Cat \\
30 & What are the 2026 recommendations for maropitant use as a cough suppressant in dogs with tracheal collapse? & Respiratory & Treatment (agent) & Dog \\
31 & What does the 2026 evidence show regarding maropitant's anti-inflammatory properties in avian respiratory disease? & Respiratory & Treatment (agent) & Birds \\
32 & What are the most recent findings regarding biomarkers that distinguish MUE from neoplasia in dogs? & Neurology & Diagnostic strategy & Dog \\
33 & What is the most recent evidence regarding immunotherapy for lymphoma in dogs? & Oncology & Treatment (agent) & Dog \\
34 & What is the current evidence regarding LOPP versus CHOP as first-line treatment for canine T-cell lymphoma? & Oncology & Treatment (agent) & Dog \\
35 & What are the most recent findings regarding chemotherapy regimens for matrical carcinoma in dogs? & Oncology & Treatment (agent) & Dog \\
36 & What is the current evidence regarding levetiracetam dosage for seizures following surgery? & Neurology & Dosing and pharmacokinetics & N/A \\
37 & What is the current evidence regarding the effectiveness of prophylactic fenestration in preventing recurrence of cervical IVDD? & Neurology & Prevention and prophylaxis & N/A \\
38 & What is the current evidence for MRI-guided stereotactic radiation therapy as a non-surgical treatment option for spinal cord compression in cats? & Neurology & Treatment (strategy) & Cat \\
39 & What is the current evidence regarding MCT oil dosing in dogs with epilepsy? & Neurology & Dosing and pharmacokinetics & Dog \\
40 & What are the 2026 recommendations for treating presumed structural epilepsy in geriatric cats? & Neurology & Treatment (agent) & Cat \\
41 & What is the current evidence regarding the use of DOACs for PTE prevention in dogs? & Cardiology & Prevention and prophylaxis & Dog \\
42 & What does current evidence show regarding torsemide compared to furosemide in dogs with CHF? & Cardiology & Treatment (agent) & Dog \\
43 & What is the current evidence for combining ivabradine with beta-blockers in the treatment of refractory SVT? & Cardiology & Treatment (agent) & N/A \\
44 & What is the current evidence regarding the effectiveness of oral sotalol for long-term management of ectopic SVT? & Cardiology & Treatment (agent) & N/A \\
45 & What is the current evidence for managing hypotension caused by isoflurane in dogs? & Anesthesia and analgesia & Treatment (agent) & Dog \\
46 & What is the current evidence regarding the usefulness of free cortisol measurements for CIRCI? & Emergency and critical care & Diagnostic strategy & N/A \\
47 & What are the 2026 recommendations for fluid therapy in cats with hemorrhagic shock? & Emergency and critical care & Treatment (agent) & Cat \\
48 & Create a comprehensive protocol using current research literature and published articles for managing canine and feline patients with hypoglycemia. & Emergency and critical care & Protocol synthesis & Dog+cat \\
49 & What does current evidence show regarding the effectiveness of gabapentin compared to amitriptyline for managing chronic FIC in cats? & Nephrology and urology & Treatment (agent) & Cat \\
50 & What are the 2026 recommendations for using frunevetmab in cats with idiopathic cystitis? & Nephrology and urology & Treatment (agent) & Cat \\
51 & What empirical antibiotics are recommended in 2026 for canine E. coli cystitis while awaiting culture results? & Nephrology and urology & Treatment (agent) & Dog \\
52 & What is the current evidence regarding the efficacy rates of percutaneous compared to surgical tube cystostomy in goats? & Nephrology and urology & Treatment (strategy) & Goat \\
53 & What is the current evidence for using low-dose ketamine in horses with abdominal pain? & Anesthesia and analgesia & Treatment (agent) & Horse \\
54 & What does current evidence show regarding the safety of long-term NSAID use in cats with osteoarthritis? & Clinical pharmacology & Drug safety & Cat \\
55 & What is the current evidence regarding the effectiveness of monoclonal antibody treatments such as lokivetmab for Pyoderma in German Shepherds? & Dermatology & Treatment (agent) & Dog \\
56 & What does current evidence show regarding the length of antibiotic treatment for canine deep pyoderma? & Dermatology & Dosing and pharmacokinetics & Dog \\
57 & Recent studies on raisin toxicity in dogs & Toxicology & Etiology update & Dog \\
58 & What is the current evidence on the effectiveness of ILE for toxicities from non-lipophilic drugs? & Toxicology & Treatment (agent) & N/A \\
59 & What is the current evidence for trazodone use in pregnant dogs with anxiety before cesarean section? & Behavior & Drug safety & Dog \\
60 & What is the current evidence for using atypical antipsychotics in dogs with compulsive disorders? & Behavior & Treatment (agent) & Dog \\
61 & What are the safety protocols for managing cats following Zorbium administration in 2026? & Anesthesia and analgesia & Drug safety & Cat \\
62 & What does current evidence show regarding conservative versus surgical treatment for early lateral digital extensor tendon luxation in dogs? & Orthopedics & Treatment (strategy) & Dog \\
63 & What is the current evidence comparing surgical versus medical treatment for corneal sequestrum in cats? & Ophthalmology & Treatment (strategy) & Cat \\
64 & What does current evidence show regarding EHV-1 vaccine effectiveness against neurological disease? & Infectious disease & Prevention and prophylaxis & Horse \\
65 & What is the current evidence regarding vacuum-assisted closure compared to active suction drains? & Surgery & Treatment (strategy) & N/A \\
66 & What do the 2026 guidelines recommend regarding medical versus surgical treatment of equine omphalophlebitis? & Surgery & Treatment (strategy) & Horse \\
67 & What is the most recent evidence for the treatment of equine neutropenic stomatitis? & Dentistry and oral medicine & Treatment (agent) & Horse \\
\end{longtable}
\end{small}
\twocolumn

\begin{table*}[t]
\centering
\small
\begin{tabular}{lcccl}
\toprule
\textbf{Name} & \textbf{Precision} & \textbf{Temperature} & \textbf{Reasoning effort} & \textbf{Tag} \\
\midrule
Claude Opus 4.6 & -- & 1.0 & high & \texttt{anthropic/claude-opus-4.6} \\
Gemini 3.1 Pro & -- & 1.0 & high & \texttt{google/gemini-3.1-pro-preview} \\
GPT-5.4 & -- & 1.0 & high & \texttt{openai/gpt-5.4} \\
DeepSeek V3.2 & \texttt{fp8} & 1.0 & -- & \texttt{deepseek/deepseek-v3.2} \\
Qwen3.5 397B & \texttt{fp8} & 1.0 & -- & \texttt{qwen/qwen3.5-397b-a17b} \\
Mistral Small 4 & \texttt{fp8} & 0.6 & -- & \texttt{mistralai/mistral-small-2603} \\
\bottomrule
\end{tabular}
\caption{Models and the corresponding OpenRouter tags used to generate the outputs in our dataset.}
\label{tab:output_models}
\end{table*}

\clearpage
\onecolumn
\begin{small}
\begin{longtable}{p{4.5cm}p{4.75cm}p{4.75cm}}
\caption{Error analysis of claim decomposition.}
\label{tab:decomp_errors} \\
\toprule
\textbf{Segment} & \textbf{Claims} & \textbf{Issue} \\
\midrule
\endfirsthead

\multicolumn{3}{l}%
{\emph{Table \ref{tab:decomp_errors} -- continued from previous page}} \\
\toprule
\textbf{Segment} & \textbf{Claims} & \textbf{Issue} \\
\midrule
\endhead

\midrule
\multicolumn{3}{r}{\emph{Continued on next page}} \\
\endfoot

\bottomrule
\endlastfoot

**Exaggerated Dosages**: Studies administering robenacoxib at extreme dosages---such as 20 times the recommended dose for 42 days (10 mg/kg q12h)---resulted in no observable toxic effects.
& 1. Studies have evaluated the administration of robenacoxib at extreme dosages. \newline 
2. One extreme dosage studied was 20 times the recommended dose of robenacoxib. \newline
3. The duration of the extreme dosage study was 42 days. \newline
\textcolor{red}{\textbf{4. The extreme dosage was administered at 10 mg/kg every 12 hours (q12h).}} \newline 5. The administration of robenacoxib at 20 times the recommended dose for 42 days resulted in no observable toxic effects. &
\emph{Suboptimal decomposition} \newline Claim (4) contains both a dose and a frequency claims. These should be decomposed into two separate claims for the system to be able to verify dosage and frequency separately. \\
\midrule

In a prospective, randomized, double-blinded clinical study (n=16 female dogs undergoing ovariectomy), trazodone (5 mg/kg orally 2 hours prior to hospital arrival) significantly reduced stress scores on the Clinic Dog Stress Scale (CDSS) compared to placebo (mean score 5.12 $\pm$ 2.95 vs 11.50 $\pm$ 6.16; \emph{p} = 0.02), and reduced required doses of propofol for induction (1.36 $\pm$ 0.55 mg/kg vs 2.13 $\pm$ 1.20 mg/kg), though not statistically significant, it showed a trend toward lower doses in the trazodone group. & \textcolor{red}{\textbf{1. A prospective, randomized, double-blinded clinical study was conducted involving 16 female dogs undergoing ovariectomy.}} \newline
2. Trazodone was administered at a dose of 5 mg/kg orally. \newline
3. Trazodone was administered 2 hours prior to hospital arrival. \newline
4. Trazodone significantly reduced stress scores on the Clinic Dog Stress Scale (CDSS) compared to placebo. \newline
5. The mean CDSS score for the trazodone group was 5.12 $\pm$ 2.95. \newline
6. The mean CDSS score for the placebo group was 11.50 $\pm$ 6.16. \newline
7. The p-value for the difference in CDSS scores was 0.02. \newline
8. Trazodone reduced the required doses of propofol for induction compared to placebo. \newline
9. The required dose of propofol for induction in the trazodone group was 1.36 $\pm$ 0.55 mg/kg. \newline
10. The required dose of propofol for induction in the placebo group was 2.13 $\pm$ 1.20 mg/kg. \newline
11. The reduction in propofol dose for induction was not statistically significant. \newline
12. The study results showed a trend toward lower propofol doses in the trazodone group. & \emph{Suboptimal decomposition} \newline The attributes of the study (prospective, randomized, double-blinded, involving 16 dogs) should each have a corresponding atomic claim. \\
\midrule

Efficacy has been demonstrated in trials evaluating diets ``enriched with 6.5 per cent MCT oil (as fed basis)''. & \textcolor{red}{\textbf{1. Efficacy has been demonstrated in trials evaluating diets enriched with MCT oil.}} \newline 2. The diets evaluated in these trials were enriched with 6.5\% MCT oil. \newline 3. The 6.5\% MCT oil enrichment level is calculated on an as-fed basis. & \emph{Suboptimal decomposition} \newline The claim could be further decomposed to two separate claims: one about the trials and the other about their efficacy. \\

Further research is required, as the literature explicitly concludes that ``studies investigating the pharmacokinetics of this medication in other species\ldots and performance of pharmacodynamic studies evaluating the potential for use as an\ldots anti-inflammatory\ldots drug in birds are warranted'' &
1. Further research on maropitant is required. \newline
2. The literature concludes that studies investigating the pharmacokinetics of maropitant in other species are warranted. \newline
\textcolor{red}{\textbf{3. The literature concludes that pharmacodynamic studies evaluating the potential for maropitant's use as an anti-inflammatory drug in birds are warranted.}} \newline
\textcolor{red}{\textbf{4. The literature concludes that the performance of pharmacodynamic studies for maropitant in birds is warranted.}} &
\emph{Semantic drift} \newline Coordination scope error. \\
\midrule
The combination of advanced age and a structural brain lesion signifies a higher risk of early mortality. & \textcolor{red}{\textbf{1. Advanced age signifies a higher risk of early mortality.}} \newline \textcolor{red}{\textbf{2. A structural brain lesion signifies a higher risk of early mortality.}} \newline 3. The combination of advanced age and a structural brain lesion signifies a higher risk of early mortality. & \emph{Semantic drift} \newline Fallacy of division: the original claim links the risk to the combination of the two factors, not to any of the factors in isolation. \\
\midrule

Dogs with no antimicrobials had a median hospitalization of 1 day and showed significant improvement after fluid resuscitation, underscoring that supportive care is sufficient in most cases. & 1. Dogs that received no antimicrobials had a median hospitalization of 1 day. \newline 2. Dogs that received no antimicrobials showed significant improvement after fluid resuscitation. \newline 3. Supportive care is sufficient in most cases of AHDS. & \emph{Semantic drift} \newline Missing evidential link between the first two claims and the last claim. \\
\midrule

The **most defensible evidence-based position** is that ponazuril is a biologically credible option because it is toltrazuril’s main metabolite and has pharmacokinetic features compatible with durable anticoccidial activity, while toltrazuril itself has broad evidence of efficacy in other species but also known limitations related to timing and possible resistance. & 1. The most defensible evidence-based position is that ponazuril is a biologically credible option. \newline 2. Ponazuril is the main metabolite of toltrazuril. \newline 3. Ponazuril has pharmacokinetic features compatible with durable anticoccidial activity. \newline 4. Toltrazuril has broad evidence of efficacy in species other than dogs. \newline 5. Toltrazuril has known limitations related to the timing of treatment. \newline 6. Toltrazuril has known limitations related to possible resistance. & \emph{Semantic drift} \newline Missing causal link between claim (1) and claims (2) and (3). \\
\midrule

The current evidence supporting the use of low-dose ketamine specifically for the medical management of abdominal pain (colic) in horses is limited, as non-steroidal anti-inflammatory drugs (NSAIDs) remain the standard of care. & \textcolor{red}{\textbf{1. Low-dose ketamine is used for the medical management of abdominal pain in horses.}} \newline 2. Abdominal pain in horses is also known as colic. \newline 3. The current evidence supporting the use of low-dose ketamine for the medical management of abdominal pain in horses is limited. \newline 4. Non-steroidal anti-inflammatory drugs (NSAIDs) are the standard of care for the medical management of abdominal pain in horses. & \emph{Semantic drift} \newline A presupposition is incorrectly interpreted as an entailment: ``current evidence supporting the use of low-dose ketamine is limited'' does not entail that ketamine is \emph{actually used} in that context. \\

The available evidence must be extrapolated from studies in non-pregnant dogs undergoing surgery, which suggest trazodone can effectively reduce pre-surgical stress and anesthetic requirements. & \textcolor{red}{\textbf{1. Evidence regarding trazodone use in this scenario must be extrapolated from studies in non-pregnant dogs undergoing surgery.}} \newline 2. Studies in non-pregnant dogs undergoing surgery suggest trazodone can reduce pre-surgical stress. \newline 3. Studies in non-pregnant dogs undergoing surgery suggest trazodone can reduce anesthetic requirements. & \emph{Contextualization} \newline The anaphora (``this scenario'') is not resolved properly.\\
\midrule
The same cohort reported that ivabradine was discontinued in 4 of 7 SVT patients after resolution of tachyarrhythmia and improvement in heart rate, supporting short- to medium-term effectiveness in selected infants, although not specifically proving superiority of the beta-blocker combination over other regimens. & \textcolor{red}{\textbf{1. The same cohort reported that ivabradine was discontinued in 4 of 7 SVT patients.}} \newline 2. The discontinuation of ivabradine occurred after resolution of tachyarrhythmia. \newline 3. The discontinuation of ivabradine occurred after improvement in heart rate. \newline 4. The findings support the short-term effectiveness of ivabradine in selected infants. \newline 5. The findings support the medium-term effectiveness of ivabradine in selected infants. \newline 6. The findings do not specifically prove superiority of the beta-blocker combination over other regimens. & \emph{Contextualization} \newline The anaphora (``the same cohort'') is not resolved properly. \\
\midrule

As of 2026, the most current and high-certainty evidence indicates that **continuous or prolonged infusion of beta-lactam antibiotics is recommended for dogs with sepsis when pharmacodynamic targets (time above MIC) are difficult to achieve with intermit dosing**. &
\textcolor{red}{\textbf{1. As of 2026, the evidence for beta-lactam infusion in dogs with sepsis is the most current.}} \newline
\textcolor{red}{\textbf{2. The evidence supporting the recommendations is of high certainty.}} \newline
3. Continuous infusion of beta-lactam antibiotics is recommended for dogs with sepsis. \newline
4. Prolonged infusion of beta-lactam antibiotics is recommended for dogs with sepsis. \newline
5. The recommendation for continuous or prolonged infusion applies when pharmacodynamic targets are difficult to achieve with intermittent dosing. \newline
6. The pharmacodynamic target for beta-lactam antibiotics is time above MIC (minimum inhibitory concentration). &
\emph{Uninformative} \newline The claims are not informative, since the first claim is tautological, while the second claim is unverifiable. \\

\end{longtable}
\end{small}
\twocolumn

\begin{table*}[t]
\centering
\small
\begin{tabular}{cccc}
\toprule
\# & Original & Fixed & $\Delta$ \\
\midrule
1 & 1.000 & 0.733 & -0.267 \\
2 & 0.469 & 0.331 & -0.138 \\
3 & 0.312 & 0.444 & 0.132 \\
4 & 0.385 & 0.286 & -0.099 \\
5 & 0.310 & 0.261 & -0.049 \\
6 & 0.429 & 0.451 & 0.022 \\
7 & 0.317 & 0.306 & -0.011 \\
8 & 1.000 & 1.000 & 0.000 \\
9 & 0.000 & 0.000 & 0.000 \\
10 & 1.000 & 1.000 & 0.000 \\
11 & 1.000 & 1.000 & 0.000 \\
\bottomrule
\end{tabular}
\caption{Comparison of combined scores obtained from running the rest of the pipeline on the original 11 segments with identified issues (see Table~\ref{tab:decomp_errors}) and the same segments where these issues are manually corrected. $\Delta$ shows the difference of scores for each segment.}
\label{tab:decomp_score_deltas}
\end{table*}

\begin{table*}[t]
\centering
\small
\begin{tabular}{lcl}
\toprule
\textbf{Name} & \textbf{Precision} & \textbf{Tag} \\
\midrule
\multicolumn{3}{l}{\textit{Google API}} \\
\noalign{\vskip 1pt}
Gemini 3.1 Pro   & --            & \texttt{gemini-3.1-pro-preview} \\
Gemini 3 Flash   & --            & \texttt{gemini-3-flash-preview} \\
Gemini 3.6 Flash & --            & \texttt{gemini-3.6-flash} \\
\noalign{\vskip 2pt}
\hdashline[0.5pt/1.5pt]
\noalign{\vskip 3pt}
\multicolumn{3}{l}{\textit{OpenRouter}} \\
\noalign{\vskip 1pt}
GPT-5.6 Sol      & --            & \texttt{openai/gpt-5.6-sol} \\
Claude Sonnet 5  & --            & \texttt{anthropic/claude-sonnet-5} \\
Qwen3.5 35B A3B  & \texttt{fp8}  & \texttt{qwen/qwen3.5-35b-a3b} \\
Gemma 4 31B      & \texttt{bf16} & \texttt{google/gemma-4-31b-it} \\
GLM-4.7 Flash    & \texttt{bf16} & \texttt{z-ai/glm-4.7-flash} \\
Nemotron 3 Nano 30B A3B & \texttt{fp8} & \texttt{nvidia/nemotron-3-nano-30b-a3b} \\
\noalign{\vskip 2pt}
\hdashline[0.5pt/1.5pt]
\noalign{\vskip 3pt}
\multicolumn{3}{l}{\textit{Ollama}} \\
\noalign{\vskip 1pt}
Baichuan-M3 235B & \texttt{q4\_k\_m} & \texttt{baichuan-inc/baichuan-m3-235b:q4\_k\_m} \\
MedGemma 27B     & \texttt{bf16} & \texttt{medgemma:27b-it-bf16} \\
MedGemma 1.5 4B  & \texttt{bf16} & \texttt{medgemma1.5:4b-it-bf16} \\
\bottomrule
\end{tabular}
\caption{Judge models used in our experiments. The Gemini models were run via the Google API. Medical LLMs were run locally through the Ollama platform. The remaining models were run through the OpenRouter API (with a single fixed provider). \textit{Precision} is the served quantization.}
\label{tab:judge_models}
\end{table*}

\begin{table*}[t]
    \centering
    \small
    \setlength{\tabcolsep}{6pt}
    \begin{tabular}{lccccc}
    \toprule
    \textbf{Model} & \shortstack{\textbf{Fact}\\ \textbf{verification}} & \shortstack{\textbf{Harm}\\ \textbf{potential}} & \shortstack{\textbf{Combined}\\\textbf{(local)}} & \shortstack{\textbf{Combined}\\\textbf{(global)}} & \shortstack{\textbf{Combined}\\\textbf{(absolute)}} \\
    \midrule
    \multicolumn{6}{l}{\textsc{VetScore}} \\
    \noalign{\vskip 1pt}
    GPT-5.6 Sol        & $0.758_{\pm 0.004}$ & $0.713_{\pm 0.003}$ & $0.738_{\pm 0.003}$ & $0.779_{\pm 0.006}$ & $0.732_{\pm 0.005}$ \\
    Claude Sonnet 5    & $0.707_{\pm 0.005}$ & $0.689_{\pm 0.003}$ & $0.688_{\pm 0.008}$ & $0.693_{\pm 0.012}$ & $0.663_{\pm 0.011}$ \\
    Gemini 3.1 Pro (low) & $0.781_{\pm 0.004}$ & $0.761_{\pm 0.003}$ & $0.757_{\pm 0.006}$ & $0.778_{\pm 0.006}$ & $0.707_{\pm 0.004}$ \\
    Gemini 3 Flash     & $0.678_{\pm 0.006}$ & $0.711_{\pm 0.007}$ & $0.637_{\pm 0.010}$ & $0.671_{\pm 0.012}$ & $0.614_{\pm 0.007}$ \\
    (low)              & $0.697_{\pm 0.003}$ & $0.724_{\pm 0.005}$ & $0.647_{\pm 0.007}$ & $0.660_{\pm 0.014}$ & $0.613_{\pm 0.005}$ \\
    Gemini 3.6 Flash   & $0.800_{\pm 0.002}$ & $0.732_{\pm 0.003}$ & $0.776_{\pm 0.004}$ & $0.798_{\pm 0.005}$ & $0.750_{\pm 0.005}$ \\
    (low)              & $0.794_{\pm 0.002}$ & $0.737_{\pm 0.002}$ & $0.773_{\pm 0.003}$ & $0.797_{\pm 0.013}$ & $0.748_{\pm 0.003}$ \\
    \noalign{\vskip 2pt}
    \hdashline[0.5pt/1.5pt]
    \noalign{\vskip 3pt}
    Gemma 4 31B        & $0.739_{\pm 0.002}$ & $0.719_{\pm 0.003}$ & $0.716_{\pm 0.003}$ & $0.731_{\pm 0.006}$ & $0.684_{\pm 0.003}$ \\
    Qwen3.5 35B A3B    & $0.702_{\pm 0.008}$ & $0.677_{\pm 0.004}$ & $0.678_{\pm 0.014}$ & $0.698_{\pm 0.020}$ & $0.648_{\pm 0.010}$ \\
    GLM-4.7 Flash 30B  & $0.456_{\pm 0.006}$ & $0.573_{\pm 0.014}$ & $0.419_{\pm 0.013}$ & $0.472_{\pm 0.018}$ & $0.418_{\pm 0.015}$ \\
    Nemotron 3 30B A3B & $0.411_{\pm 0.012}$ & $0.547_{\pm 0.013}$ & $0.403_{\pm 0.019}$ & $0.432_{\pm 0.039}$ & $0.407_{\pm 0.021}$ \\
    \noalign{\vskip 2pt}
    \hdashline[0.5pt/1.5pt]
    \noalign{\vskip 3pt}
    Baichuan-M3 235B   & $0.561_{\pm 0.011}$ & $0.678_{\pm 0.011}$ & $0.547_{\pm 0.012}$ & $0.592_{\pm 0.024}$ & $0.559_{\pm 0.019}$ \\
    MedGemma 27B       & $0.446_{\pm 0.009}$ & $0.660_{\pm 0.008}$ & $0.415_{\pm 0.006}$ & $0.444_{\pm 0.027}$ & $0.395_{\pm 0.015}$ \\
    MedGemma 1.5 4B        & $0.361_{\pm 0.014}$ & $0.465_{\pm 0.017}$ & $0.332_{\pm 0.015}$ & $0.382_{\pm 0.017}$ & $0.319_{\pm 0.008}$ \\
    \midrule
    \multicolumn{6}{l}{\textit{Verification-only}} \\
    \noalign{\vskip 1pt}
    GPT-5.6 Sol        & $0.758_{\pm 0.004}$ & - & $0.743_{\pm 0.001}$ & $0.776_{\pm 0.004}$ & $0.669^{***}_{\pm 0.004}$ \\
    Claude Sonnet 5    & $0.707_{\pm 0.005}$ & - & $0.697_{\pm 0.007}$ & $0.698_{\pm 0.010}$ & $0.615^{***}_{\pm 0.009}$ \\
    Gemini 3.1 Pro (low) & $0.781_{\pm 0.004}$ & - & $0.759_{\pm 0.006}$ & $0.774_{\pm 0.003}$ & $0.645^{***}_{\pm 0.003}$ \\
    Gemini 3 Flash     & $0.678_{\pm 0.006}$ & - & $0.641_{\pm 0.009}$ & $0.674_{\pm 0.015}$ & $0.568^{***}_{\pm 0.007}$ \\
    (low)              & $0.697_{\pm 0.003}$ & - & $0.653_{\pm 0.007}$ & $0.665_{\pm 0.012}$ & $0.565^{***}_{\pm 0.004}$ \\
    Gemini 3.6 Flash   & $0.800_{\pm 0.002}$ & - & $0.787_{\pm 0.003}$ & $0.814_{\pm 0.005}$ & $0.683^{***}_{\pm 0.007}$ \\
    (low)              & $0.794_{\pm 0.002}$ & - & $0.783^{*}_{\pm 0.004}$ & $0.807_{\pm 0.009}$ & $0.683^{***}_{\pm 0.006}$ \\
    \noalign{\vskip 2pt}
    \hdashline[0.5pt/1.5pt]
    \noalign{\vskip 3pt}
    Gemma 4 31B        & $0.739_{\pm 0.002}$ & - & $0.725_{\pm 0.002}$ & $0.742_{\pm 0.006}$ & $0.632^{***}_{\pm 0.001}$ \\
    Qwen3.5 35B A3B    & $0.702_{\pm 0.008}$ & - & $0.682_{\pm 0.011}$ & $0.713_{\pm 0.016}$ & $0.588^{***}_{\pm 0.010}$ \\
    GLM-4.7 Flash 30B  & $0.456_{\pm 0.006}$ & - & $0.418_{\pm 0.012}$ & $0.478_{\pm 0.016}$ & $0.399^{***}_{\pm 0.016}$ \\
    Nemotron 3 30B A3B & $0.411_{\pm 0.012}$ & - & $0.402_{\pm 0.018}$ & $0.446_{\pm 0.037}$ & $0.391^{***}_{\pm 0.020}$ \\
    \noalign{\vskip 2pt}
    \hdashline[0.5pt/1.5pt]
    \noalign{\vskip 3pt}
    Baichuan-M3 235B   & $0.561_{\pm 0.011}$ & - & $0.540_{\pm 0.013}$ & $0.574_{\pm 0.028}$ & $0.480^{***}_{\pm 0.020}$ \\
    MedGemma 27B       & $0.446_{\pm 0.009}$ & - & $0.407^{*}_{\pm 0.005}$ & $0.446_{\pm 0.020}$ & $0.355^{***}_{\pm 0.013}$ \\
    MedGemma 1.5 4B        & $0.361_{\pm 0.014}$ & - & $0.330_{\pm 0.014}$ & $0.390_{\pm 0.023}$ & $0.301^{***}_{\pm 0.007}$ \\
    \midrule
    \multicolumn{6}{l}{\textit{End-to-end}} \\
    \noalign{\vskip 2pt}
    GPT-5.6 Sol        & - & - & $0.319^{***}_{\pm 0.004}$ & $0.289^{***}_{\pm 0.013}$ & - \\
    Gemini 3.6 Flash   & - & - & $0.452^{***}_{\pm 0.012}$ & $0.454^{***}_{\pm 0.008}$ & - \\
    \noalign{\vskip 2pt}
    \hdashline[0.5pt/1.5pt]
    \noalign{\vskip 3pt}
    Gemma 4 31B        & - & - & $0.485^{***}_{\pm 0.014}$ & $0.464^{***}_{\pm 0.013}$ & - \\
    Qwen3.5 35B A3B    & - & - & $0.313^{***}_{\pm 0.015}$ & $0.305^{***}_{\pm 0.036}$ & - \\
    \bottomrule
    \end{tabular}
    \caption{Spearman ($\rho$) correlations with human annotations for the full \textsc{VetScore} pipeline and two baselines. \textit{Verification-only} drops the harm-potential component and scores each segment by fact verification alone. \textit{End-to-end} produces the combined score in a single call, without decomposition. Each \textit{Combined} column correlates the setup's score with the human combined score, normalized \textit{locally} (per segment) or \textit{globally} (per output), or in the non-normalized \textit{absolute} form. Each cell is the average over $n=5$ runs with the standard deviation as a subscript. In the \textit{Verification-only} and \textit{End-to-end} sections, asterisks indicate a significant difference of the \textit{Combined} score from the same model's \textsc{VetScore} value (paired bootstrap over segments, 10,000 resamples: $^{*}p<0.05$, $^{**}p<0.01$, $^{***}p<0.001$).}
    \label{tab:full_results}
    \end{table*}

\begin{table*}[t]
\centering
\small
\setlength{\tabcolsep}{5pt}
\begin{tabular}{lrrrrrccc}
\toprule
 & & & \multicolumn{3}{c}{\textbf{Reason (\% of FP)}} & \multicolumn{3}{c}{\textbf{Score}} \\
\cmidrule(lr){4-6} \cmidrule(lr){7-9}
\textbf{Judge} & \textbf{\#} & \textbf{FP} & \textbf{Support} & \textbf{Harm} & \textbf{Both} & \textbf{Precision} & \textbf{Recall} & \textbf{F1} \\
\midrule
GPT-5.6 Sol            & 417 & 150 & 9\% & 74\% & 17\% & 0.64 & 0.82 & 0.72 \\
Claude Sonnet 5        & 308 & 93 & 9\% & 82\% & 9\% & 0.70 & 0.66 & 0.68 \\
Gemini 3.1 Pro (low)   & 297 & 73 & 7\% & 83\% & 10\% & 0.75 & 0.69 & 0.72 \\
Gemini 3 Flash         & 307 & 98 & 7\% & 87\% & 6\% & 0.68 & 0.65 & 0.66 \\
Gemini 3.6 Flash       & 401 & 141 & 12\% & 72\% & 16\% & 0.65 & 0.80 & 0.72 \\
\noalign{\vskip 2pt}
\hdashline[0.5pt/1.5pt]
\noalign{\vskip 3pt}
Gemma 4 31B            & 307 & 89 & 8\% & 81\% & 11\% & 0.71 & 0.67 & 0.69 \\
Qwen3.5 35B A3B        & 314 & 105 & 12\% & 74\% & 14\% & 0.67 & 0.65 & 0.66 \\
GLM-4.7 Flash 30B      & 189 & 63 & 6\% & 85\% & 9\% & 0.67 & 0.39 & 0.49 \\
Nemotron 3 30B A3B     & 215 & 84 & 16\% & 68\% & 16\% & 0.61 & 0.40 & 0.49 \\
\noalign{\vskip 2pt}
\hdashline[0.5pt/1.5pt]
\noalign{\vskip 3pt}
Baichuan-M3 235B       & 246 & 75 & 24\% & 58\% & 18\% & 0.70 & 0.53 & 0.60 \\
MedGemma 27B           & 153 & 44 & 6\% & 82\% & 12\% & 0.71 & 0.34 & 0.46 \\
MedGemma 1.5 4B        & 164 & 68 & 11\% & 67\% & 22\% & 0.59 & 0.30 & 0.39 \\
\bottomrule
\end{tabular}
\caption{False positives of high-risk segment identification. A segment is flagged if it contains a claim that is unsupported and high-harm. For human experts, a majority of annotators judge it unsupported and its MFRM fair harm score is $\geq 3.43$. For a judge, the sample verdict is unsupported and the harm score is $\geq 4$. \textit{\#} = number of segments flagged by the judge; \textit{FP} = false positives. The \textit{Reason} columns indicate the reason for claims not being identified as high-risk by human experts: \textit{Support} = the majority of annotators judge them as supported; \textit{Harm} = the annotators agree that the claims are unsupported, but their harm score is below 3.43; \textit{Both} = both reasons, or different reasons for different claims.}
\label{tab:risk_identification}
\end{table*}

\begin{table*}[t]
\centering
\small
\setlength{\tabcolsep}{4pt}
\begin{tabular}{llccc}
\toprule
\textbf{Model} & \textbf{Decomp.} & \shortstack{\textbf{Combined}\\\textbf{(local)}} & \shortstack{\textbf{Combined}\\\textbf{(global)}} & \shortstack{\textbf{Combined}\\\textbf{(absolute)}} \\
\midrule
Gemini 3.6 Flash & Original & $0.776_{\pm 0.004}$ & $0.798_{\pm 0.005}$ & $0.750_{\pm 0.005}$ \\
                 & Self  & $0.715_{\pm 0.007}$ & $0.746_{\pm 0.013}$ & $0.700_{\pm 0.007}$ \\
\midrule
Gemma 4 31B      & Original & $0.716_{\pm 0.003}$ & $0.731_{\pm 0.006}$ & $0.684_{\pm 0.003}$ \\
                 & Self  & $0.665_{\pm 0.003}$ & $0.660_{\pm 0.003}$ & $0.647_{\pm 0.002}$ \\
\bottomrule
\end{tabular}
\caption{Spearman correlations of the combined score with human annotations when the judge performs the claim decomposition itself (\textit{Self}) versus using the shared decomposition rated by the annotators (\textit{Original}). Since self-decomposed claims are different from the annotated ones, the combined score is aggregated per segment (\textit{local}, \textit{absolute}) or per output (\textit{global}) on both sides. Values are the mean $\pm$ standard deviation over five samples.}
\label{tab:selfdecomp_results}
\end{table*}

\begin{table*}[t]
    \centering
    \small
    \setlength{\tabcolsep}{6pt}
    \begin{tabular}{lcccccccccc}
    \toprule
    \textbf{Model} & \multicolumn{2}{c}{$p=1$} & \multicolumn{2}{c}{$p=2$} & \multicolumn{2}{c}{$p=2.426$} & \multicolumn{2}{c}{$p=3$} & \multicolumn{2}{c}{max. $\Delta$} \\
    \cmidrule(lr){2-3} \cmidrule(lr){4-5} \cmidrule(lr){6-7} \cmidrule(lr){8-9} \cmidrule(lr){10-11}
     & $\rho$ & rank & $\rho$ & rank & $\rho$ & rank & $\rho$ & rank & $\rho$ & rank \\
    \midrule
    \multicolumn{11}{l}{\textit{Combined (local)}} \\
    \noalign{\vskip 1pt}
    Gemini 3.6 Flash     & $0.792$ & 1 & $0.779$ & 1 & $0.776$ & 1 & $0.773$ & 1 & $0.020$ & 0 \\
    Gemini 3.1 Pro (low) & $0.771$ & 2 & $0.760$ & 2 & $0.757$ & 2 & $0.753$ & 2 & $0.018$ & 0 \\
    GPT-5.6 Sol          & $0.750$ & 3 & $0.740$ & 3 & $0.738$ & 3 & $0.734$ & 3 & $0.016$ & 0 \\
    Gemma 4 31B          & $0.729$ & 4 & $0.719$ & 4 & $0.716$ & 4 & $0.713$ & 4 & $0.017$ & 0 \\
    Claude Sonnet 5      & $0.700$ & 5 & $0.690$ & 5 & $0.688$ & 5 & $0.685$ & 5 & $0.015$ & 0 \\
    Qwen3.5 35B A3B      & $0.692$ & 6 & $0.681$ & 6 & $0.678$ & 6 & $0.674$ & 6 & $0.018$ & 0 \\
    Gemini 3 Flash       & $0.646$ & 7 & $0.639$ & 7 & $0.637$ & 7 & $0.635$ & 7 & $0.011$ & 0 \\
    Baichuan-M3 235B     & $0.547$ & 8 & $0.547$ & 8 & $0.547$ & 8 & $0.546$ & 8 & $0.001$ & 0 \\
    GLM-4.7 Flash 30B    & $0.421$ & 9 & $0.419$ & 9 & $0.419$ & 9 & $0.417$ & 9 & $0.005$ & 0 \\
    MedGemma 27B         & $0.417$ & 10 & $0.415$ & 10 & $0.415$ & 10 & $0.412$ & 10 & $0.005$ & 0 \\
    Nemotron 3 30B A3B   & $0.404$ & 11 & $0.403$ & 11 & $0.403$ & 11 & $0.401$ & 11 & $0.003$ & 0 \\
    MedGemma 1.5 4B      & $0.338$ & 12 & $0.334$ & 12 & $0.332$ & 12 & $0.331$ & 12 & $0.007$ & 0 \\
    \midrule
    \multicolumn{11}{l}{\textit{Combined (global)}} \\
    \noalign{\vskip 1pt}
    Gemini 3.6 Flash     & $0.832$ & 1 & $0.808$ & 1 & $0.798$ & 1 & $0.788$ & 1 & $0.044$ & 0 \\
    GPT-5.6 Sol          & $0.799$ & \textbf{3} & $0.785$ & 2 & $0.779$ & 2 & $0.771$ & 2 & $0.028$ & 1 \\
    Gemini 3.1 Pro (low) & $0.804$ & \textbf{2} & $0.785$ & 3 & $0.778$ & 3 & $0.768$ & 3 & $0.036$ & 1 \\
    Gemma 4 31B          & $0.757$ & 4 & $0.738$ & 4 & $0.731$ & 4 & $0.723$ & 4 & $0.034$ & 0 \\
    Qwen3.5 35B A3B      & $0.728$ & 5 & $0.705$ & 5 & $0.698$ & 5 & $0.688$ & 5 & $0.040$ & 0 \\
    Claude Sonnet 5      & $0.721$ & 6 & $0.701$ & 6 & $0.693$ & 6 & $0.682$ & 6 & $0.039$ & 0 \\
    Gemini 3 Flash       & $0.691$ & 7 & $0.677$ & 7 & $0.671$ & 7 & $0.664$ & 7 & $0.027$ & 0 \\
    Baichuan-M3 235B     & $0.597$ & 8 & $0.595$ & 8 & $0.592$ & 8 & $0.589$ & 8 & $0.008$ & 0 \\
    GLM-4.7 Flash 30B    & $0.475$ & 9 & $0.474$ & 9 & $0.472$ & 9 & $0.470$ & 9 & $0.005$ & 0 \\
    MedGemma 27B         & $0.457$ & 10 & $0.449$ & 10 & $0.444$ & 10 & $0.440$ & 10 & $0.017$ & 0 \\
    Nemotron 3 30B A3B   & $0.436$ & 11 & $0.434$ & 11 & $0.432$ & 11 & $0.430$ & 11 & $0.006$ & 0 \\
    MedGemma 1.5 4B      & $0.399$ & 12 & $0.387$ & 12 & $0.382$ & 12 & $0.376$ & 12 & $0.023$ & 0 \\
    \midrule
    \multicolumn{11}{l}{\textit{Combined (absolute)}} \\
    \noalign{\vskip 1pt}
    Gemini 3.6 Flash     & $0.783$ & 1 & $0.758$ & 1 & $0.750$ & 1 & $0.742$ & 1 & $0.042$ & 0 \\
    GPT-5.6 Sol          & $0.756$ & 2 & $0.740$ & 2 & $0.732$ & 2 & $0.725$ & 2 & $0.031$ & 0 \\
    Gemini 3.1 Pro (low) & $0.755$ & 3 & $0.720$ & 3 & $0.707$ & 3 & $0.695$ & 3 & $0.060$ & 0 \\
    Gemma 4 31B          & $0.721$ & 4 & $0.693$ & 4 & $0.684$ & 4 & $0.674$ & 4 & $0.047$ & 0 \\
    Claude Sonnet 5      & $0.697$ & 5 & $0.673$ & 5 & $0.663$ & 5 & $0.653$ & 5 & $0.044$ & 0 \\
    Qwen3.5 35B A3B      & $0.680$ & 6 & $0.656$ & 6 & $0.648$ & 6 & $0.639$ & 6 & $0.041$ & 0 \\
    Gemini 3 Flash       & $0.646$ & 7 & $0.623$ & 7 & $0.614$ & 7 & $0.605$ & 7 & $0.041$ & 0 \\
    Baichuan-M3 235B     & $0.560$ & 8 & $0.561$ & 8 & $0.559$ & 8 & $0.556$ & 8 & $0.005$ & 0 \\
    GLM-4.7 Flash 30B    & $0.430$ & 9 & $0.422$ & 9 & $0.418$ & 9 & $0.412$ & 9 & $0.018$ & 0 \\
    Nemotron 3 30B A3B   & $0.417$ & 10 & $0.412$ & 10 & $0.407$ & 10 & $0.401$ & 10 & $0.015$ & 0 \\
    MedGemma 27B         & $0.417$ & 11 & $0.401$ & 11 & $0.395$ & 11 & $0.386$ & 11 & $0.030$ & 0 \\
    MedGemma 1.5 4B      & $0.336$ & 12 & $0.324$ & 12 & $0.319$ & 12 & $0.312$ & 12 & $0.024$ & 0 \\
    \bottomrule
    \end{tabular}
    \caption{Effect of the weight exponent $p$ on the \textsc{VetScore} combined score. Each cell is the Spearman ($\rho$) correlation with the human combined score, averaged over $n=5$ runs, and the judge's rank at the given $p$. $p=2.426$ is the exponent used in our experiments. Ranks that differ from our main setup are bolded. \textit{max.~$\Delta$} is the largest difference between any two exponents.}
    \label{tab:exponent_sensitivity}
    \end{table*}

\begin{table*}[t]
\centering
\small
\setlength{\tabcolsep}{3pt}
\begin{tabular}{l@{\hskip 6pt}cc@{\hskip 6pt}cc@{\hskip 6pt}cc@{\hskip 6pt}cc@{\hskip 6pt}cc@{\hskip 6pt}cc}
\toprule
\textbf{System/} & \multicolumn{2}{c}{\textbf{Opus 4.6}} & \multicolumn{2}{c}{\textbf{Qwen 3.5}} & \multicolumn{2}{c}{\textbf{Gemini 3.1 Pro}} & \multicolumn{2}{c}{\textbf{GPT 5.4}} & \multicolumn{2}{c}{\textbf{DeepSeek v3.2}} & \multicolumn{2}{c}{\textbf{Mistral Small}} \\
\cmidrule(lr){2-3} \cmidrule(lr){4-5} \cmidrule(lr){6-7} \cmidrule(lr){8-9} \cmidrule(lr){10-11} \cmidrule(lr){12-13}
\textbf{Judge} & Mean & Rank & Mean & Rank & Mean & Rank & Mean & Rank & Mean & Rank & Mean & Rank \\
\midrule
\multicolumn{13}{l}{\textit{Combined}} \\
\noalign{\vskip 1pt}
Human               & 0.689 & 1--2 & 0.646 & 1--3 & 0.595 & 2--4 & 0.548 & 3--4 & 0.461 & 5--6 & 0.387 & 5--6 \\
\noalign{\vskip 2pt}
\hdashline[0.5pt/1.5pt]
\noalign{\vskip 3pt}
GPT-5.6 Sol         & 0.750 & 1--2 & 0.708 & 1--4 & 0.663 & 2--4 & 0.645 & 2--4 & 0.501 & 5 & 0.414 & 6 \\
Claude Sonnet 5     & 0.776 & 1--4 & 0.779 & 1--4 & 0.778 & 1--4 & 0.746 & 1--4 & 0.616 & 5 & 0.481 & 6 \\
Gemini 3.1 Pro      & 0.767 & 1--3 & 0.727 & 1--4 & 0.733 & 1--4 & 0.687 & 2--4 & 0.566 & 5 & 0.445 & 6 \\
Gemini 3.6 Flash    & 0.714 & 1--3 & 0.647 & 1--4 & 0.671 & 1--4 & 0.605 & 2--4 & 0.481 & 5 & 0.393 & 6 \\
Gemini 3 Flash      & 0.789 & 1--4 & 0.782 & 1--4 & 0.803 & 1--4 & 0.777 & 1--4 & 0.672 & 5 & 0.536 & 6 \\
\noalign{\vskip 2pt}
\hdashline[0.5pt/1.5pt]
\noalign{\vskip 3pt}
Gemma 4 31B         & 0.763 & 1--3 & 0.747 & 1--4 & 0.758 & 1--3 & 0.680 & 3--5 & 0.613 & 4--5 & 0.459 & 6 \\
Qwen3.5 35B-A3B     & 0.755 & 1--3 & 0.718 & 1--4 & 0.724 & 1--4 & 0.665 & 2--4 & 0.585 & 5 & 0.467 & 6 \\
GLM-4.7 Flash 30B   & 0.904 & 1--3 & 0.889 & 1--3 & 0.901 & 1--3 & 0.831 & 4--5 & 0.807 & 4--5 & 0.704 & 6 \\
Nemotron 3 30B A3B  & 0.872 & 1--4 & 0.859 & 1--4 & 0.852 & 1--4 & 0.833 & 1--5 & 0.809 & 4--6 & 0.749 & 5--6 \\
\noalign{\vskip 2pt}
\hdashline[0.5pt/1.5pt]
\noalign{\vskip 3pt}
Baichuan-M3 235B    & 0.769 & 1--3 & 0.749 & 1--4 & 0.748 & 1--4 & 0.706 & 2--4 & 0.617 & 5 & 0.513 & 6 \\
MedGemma 27B        & 0.847 & 2--5 & 0.849 & 1--4 & 0.892 & 1--2 & 0.836 & 2--5 & 0.799 & 3--5 & 0.717 & 6 \\
MedGemma 1.5 4B     & 0.885 & 1--3 & 0.871 & 1--3 & 0.891 & 1--3 & 0.787 & 4--6 & 0.817 & 4--6 & 0.759 & 4--6 \\
\midrule
\multicolumn{13}{l}{\textit{Fact verification}} \\
\noalign{\vskip 1pt}
Human               & 0.689 & 1 & 0.605 & 2--3 & 0.560 & 2--4 & 0.516 & 3--5 & 0.451 & 4--5 & 0.364 & 6 \\
\noalign{\vskip 2pt}
\hdashline[0.5pt/1.5pt]
\noalign{\vskip 3pt}
GPT-5.6 Sol         & 0.736 & 1--2 & 0.662 & 1--4 & 0.629 & 2--4 & 0.618 & 2--4 & 0.502 & 5 & 0.380 & 6 \\
Claude Sonnet 5     & 0.766 & 1--4 & 0.730 & 1--4 & 0.734 & 1--4 & 0.710 & 1--4 & 0.615 & 5 & 0.443 & 6 \\
Gemini 3.1 Pro      & 0.752 & 1 & 0.671 & 2--4 & 0.678 & 2--4 & 0.658 & 2--4 & 0.557 & 5 & 0.401 & 6 \\
Gemini 3.6 Flash    & 0.711 & 1 & 0.610 & 2--4 & 0.613 & 2--4 & 0.568 & 2--4 & 0.478 & 5 & 0.355 & 6 \\
Gemini 3 Flash      & 0.777 & 1--4 & 0.728 & 1--4 & 0.767 & 1--4 & 0.747 & 1--4 & 0.661 & 5 & 0.501 & 6 \\
\noalign{\vskip 2pt}
\hdashline[0.5pt/1.5pt]
\noalign{\vskip 3pt}
Gemma 4 31B         & 0.756 & 1--3 & 0.697 & 1--4 & 0.717 & 1--4 & 0.648 & 2--5 & 0.610 & 4--5 & 0.439 & 6 \\
Qwen3.5 35B-A3B     & 0.745 & 1--3 & 0.679 & 1--4 & 0.695 & 1--4 & 0.650 & 2--4 & 0.582 & 5 & 0.436 & 6 \\
GLM-4.7 Flash 30B   & 0.893 & 1--3 & 0.867 & 1--3 & 0.894 & 1--3 & 0.806 & 4--5 & 0.808 & 4--5 & 0.674 & 6 \\
Nemotron 3 30B A3B  & 0.875 & 1--4 & 0.838 & 1--5 & 0.853 & 1--4 & 0.841 & 1--5 & 0.808 & 3--6 & 0.753 & 5--6 \\
\noalign{\vskip 2pt}
\hdashline[0.5pt/1.5pt]
\noalign{\vskip 3pt}
Baichuan-M3 235B    & 0.739 & 1--3 & 0.718 & 1--4 & 0.709 & 1--4 & 0.662 & 2--4 & 0.575 & 5 & 0.479 & 6 \\
MedGemma 27B        & 0.815 & 1--5 & 0.826 & 1--4 & 0.858 & 1--3 & 0.796 & 2--5 & 0.772 & 3--5 & 0.690 & 6 \\
MedGemma 1.5 4B     & 0.887 & 1--3 & 0.843 & 1--4 & 0.873 & 1--3 & 0.780 & 4--6 & 0.792 & 3--6 & 0.732 & 4--6 \\
\midrule
\multicolumn{13}{l}{\textit{Harm potential}} \\
\noalign{\vskip 1pt}
Human               & 2.883 & 6 & 3.115 & 1--5 & 3.138 & 1--5 & 3.096 & 1--5 & 3.033 & 2--5 & 3.211 & 1--4 \\
\noalign{\vskip 2pt}
\hdashline[0.5pt/1.5pt]
\noalign{\vskip 3pt}
GPT-5.6 Sol         & 2.752 & 6 & 3.132 & 1--5 & 3.138 & 1--5 & 3.310 & 1--4 & 3.072 & 2--5 & 3.196 & 1--5 \\
Claude Sonnet 5     & 2.712 & 6 & 2.935 & 1--5 & 2.976 & 1--5 & 3.083 & 1--4 & 2.911 & 2--5 & 3.018 & 1--5 \\
Gemini 3.1 Pro      & 2.567 & 6 & 2.903 & 1--5 & 2.975 & 1--5 & 2.977 & 1--5 & 2.857 & 1--5 & 3.027 & 1--5 \\
Gemini 3.6 Flash    & 2.637 & 6 & 2.946 & 1--5 & 3.001 & 1--5 & 3.066 & 1--5 & 2.938 & 1--5 & 3.102 & 1--5 \\
Gemini 3 Flash      & 2.709 & 6 & 2.978 & 1--5 & 3.048 & 1--5 & 3.108 & 1--5 & 2.961 & 1--5 & 3.137 & 1--5 \\
\noalign{\vskip 2pt}
\hdashline[0.5pt/1.5pt]
\noalign{\vskip 3pt}
Gemma 4 31B         & 2.564 & 6 & 2.879 & 1--5 & 2.885 & 1--5 & 3.008 & 1--4 & 2.803 & 3--5 & 3.024 & 1--4 \\
Qwen3.5 35B-A3B     & 2.708 & 6 & 3.145 & 1--5 & 3.073 & 1--5 & 3.100 & 1--5 & 3.003 & 1--5 & 3.240 & 1--5 \\
GLM-4.7 Flash 30B   & 2.911 & 6 & 3.230 & 1--5 & 3.143 & 1--5 & 3.208 & 1--5 & 3.154 & 1--5 & 3.357 & 1--5 \\
Nemotron 3 30B A3B  & 3.190 & 1--6 & 3.344 & 1--6 & 3.298 & 1--6 & 3.257 & 1--6 & 3.209 & 1--6 & 3.381 & 1--6 \\
\noalign{\vskip 2pt}
\hdashline[0.5pt/1.5pt]
\noalign{\vskip 3pt}
Baichuan-M3 235B    & 2.617 & 5--6 & 2.897 & 1--5 & 2.887 & 1--5 & 2.903 & 1--5 & 2.764 & 2--6 & 3.041 & 1--4 \\
MedGemma 27B        & 2.715 & 6 & 3.087 & 1--5 & 3.028 & 1--5 & 3.122 & 1--5 & 2.967 & 1--5 & 3.145 & 1--5 \\
MedGemma 1.5 4B     & 3.264 & 3--6 & 3.494 & 1--4 & 3.285 & 3--6 & 3.439 & 1--6 & 3.343 & 1--6 & 3.534 & 1--4 \\
\bottomrule
\end{tabular}
\caption{Per-system scores and ranks each judge assigns to the evaluated systems (1 = highest). Scores are per-system means of the judge's five-sample average. \textit{Combined} = mean combined score; \textit{Fact verification} = mean verification rate; \textit{Harm potential} = mean harm potential (rank 1 corresponds to the most harmful content). Ranks are shown as ranges when a system's mean is not significantly different from that of the systems ranked around it (bootstrap over segments, 10,000 resamples).}
\label{tab:system_rankings}
\end{table*}

\appendix

\end{document}